\documentclass[10pt,twocolumn,letterpaper]{article}

\usepackage[pagenumbers]{style/iccv} %

\usepackage[dvipsnames]{xcolor}

\definecolor{cvprblue}{rgb}{0.21,0.49,0.74}
\usepackage[pagebackref,breaklinks,colorlinks,citecolor=cvprblue]{hyperref}

\usepackage{times}
\usepackage[dvipsnames]{xcolor}

\usepackage{multicol}
\usepackage{amsmath, amsfonts, amssymb, amsthm}
\usepackage{enumerate}
\usepackage{mathtools}
\usepackage{graphicx}
\usepackage{longtable,tabularx}
\usepackage{placeins} 
\usepackage{float}
\usepackage{multirow}
\usepackage{bbm}
\usepackage{balance}
\usepackage[ruled,algo2e]{algorithm2e}
\usepackage{algpseudocode}
\usepackage{adjustbox}
\usepackage{booktabs}
\usepackage{bm}
\usepackage{etoolbox}
\usepackage{microtype}
\usepackage{units}
\usepackage{xspace}
\usepackage[title]{appendix}
\usepackage[symbol]{footmisc}

\newcommand{\mcal}[1]{\mathcal{#1}}
\newcommand{\mbb}[1]{\mathbb{#1}}
\newcommand{\T}{\mathsf{T}}

\newcommand{\norm}[1]{\left\Vert #1 \right\Vert}

\newcommand{\algname}{\mbox{FAST-Splat}\xspace}

\newbool{extended}
\setbool{extended}{false}

\makeatletter
\newcommand{\oldnormaux}[3]{\mathpalette\oldnormaux@i{{#1}{#2}{#3}}}
\newcommand{\oldnormaux@i}[2]{\oldnormaux@ii#1#2}
\newcommand{\oldnormaux@ii}[4]{%
  \sbox\z@{$\m@th#1#2#4#3$}%
  \sbox\tw@{$\m@th\|$}%
  \mathopen{\hbox to\wd\tw@{\hss\vrule height \ht\z@ depth \dp\z@ width .2\wd\tw@\hss}}%
  #4
  \mathclose{\hbox to\wd\tw@{\hss\vrule height \ht\z@ depth \dp\z@ width .2\wd\tw@\hss}}%
}
\makeatother

\makeatletter
\newcommand{\longdash}[1][2em]{%
  \makebox[#1]{$\m@th\smash-\mkern-7mu\cleaders\hbox{$\mkern-2mu\smash-\mkern-2mu$}\hfill\mkern-7mu\smash-$}}
\makeatother
\newcommand{\omitskip}{\kern-\arraycolsep}

\def\paperID{11698} %
\def\confName{ICCV}
\def\confYear{2025}

\title{\algname: Fast, Ambiguity-Free Semantics Transfer in Gaussian Splatting}

\author{Ola Shorinwa\qquad
Jiankai Sun\qquad
Mac Schwager\\
Stanford University\\
}

\newcommand{\insertfig}{%
\includegraphics[width=\linewidth]{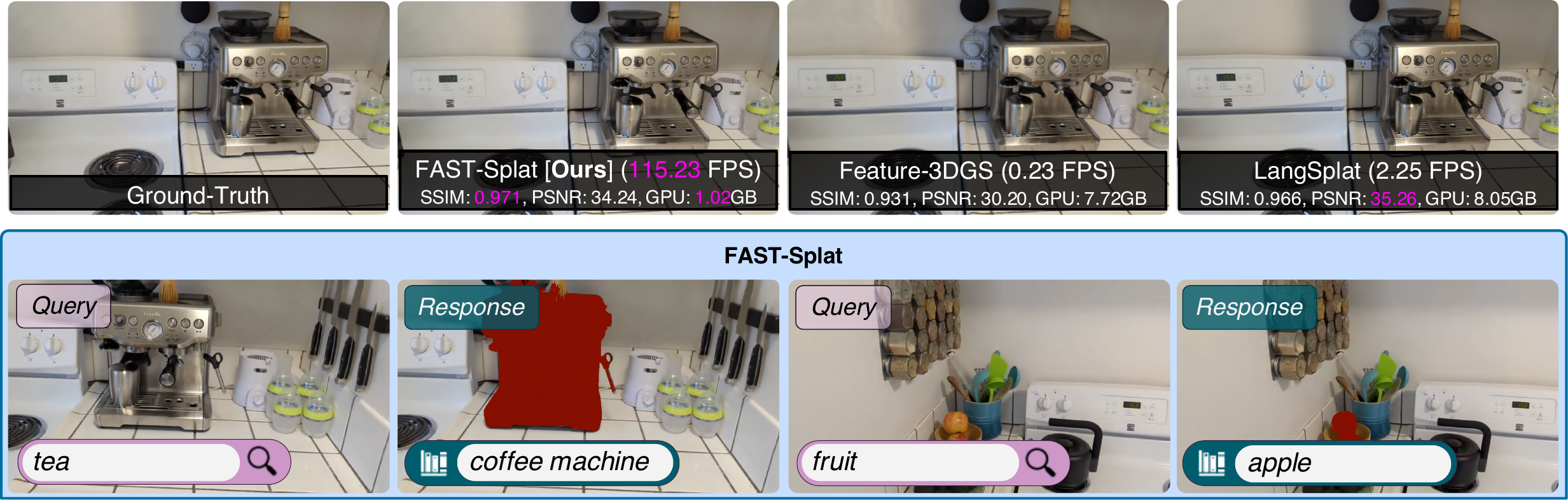}
\captionof{figure}{\algname enables fast, ambiguity-free semantic Gaussian Splatting, achieving $6$x to $8$x faster training times, $18$x to $51$x faster rendering speeds, and $6$x lower GPU memory usage compared to prior work. \algname resolves language/scene-attributed ambiguity in object localization, providing the precise semantic label of objects when given a user query, e.g., in the illustrated cases with a related but different object, \algname informs a user that it found a \emph{coffee machine} given the query: ``tea," and an \emph{apple} given the query: ``fruit."
\vspace{2ex}
}\label{fig:teaser}}

\makeatletter
\apptocmd{\@maketitle}{\centering\insertfig}{}{}%
\makeatother

\begin{document}

\maketitle
\begin{abstract}
We present \algname for fast, ambiguity-free semantic Gaussian Splatting, which seeks to address the main limitations of existing semantic Gaussian Splatting methods, namely: slow training and rendering speeds; high memory usage; and ambiguous semantic object localization. We take a bottom-up approach in deriving \algname, dismantling the limitations of closed-set semantic distillation to enable open-set (open-vocabulary) semantic distillation. Ultimately, this key approach enables \algname to provide precise semantic object localization results, even when prompted with ambiguous user-provided natural-language queries. Further, by exploiting the explicit form of the Gaussian Splatting scene representation to the fullest extent, \algname retains the remarkable training and rendering speeds of Gaussian Splatting. Precisely, while existing semantic Gaussian Splatting methods distill semantics into a separate neural field or utilize neural models for dimensionality reduction, \algname directly augments each Gaussian with specific semantic codes, preserving the training, rendering, and memory-usage advantages of Gaussian Splatting over neural field methods. These Gaussian-specific semantic codes, together with a hash-table, enable semantic similarity to be measured with open-vocabulary user prompts and further enable \algname to respond with unambiguous semantic object labels and $3$D masks, unlike prior methods. In experiments, we demonstrate that \algname is $6$\emph{x} to $8$\emph{x} faster to train, achieves between $18$\emph{x} to $51$\emph{x} faster rendering speeds, and requires about $6$\emph{x} smaller GPU memory, compared to the best-competing semantic Gaussian Splatting methods. Further, \algname achieves relatively similar or better semantic segmentation performance compared to existing methods. After the review period, we will provide links to the project website and the codebase.
\end{abstract}

\section{Introduction}
\label{sec:introduction}
Research breakthroughs in vision-language foundation models have underpinned the remarkable performance of state-of-the-art object detection and classification \cite{minderer2022simple, zhou2022detecting}, segmentation \cite{ravi2024sam, li2022language}, and image captioning \cite{radford2021learning, yu2022coca}. In general, these models learn useful multi-modal image-language representations, entirely supervised with $2$D image-text pairs, all within a shared representation space. Recent work has shown that grounding the semantic knowledge encoded by vision-language foundation models in $3$D can be useful in improving semantic segmentation and localization \cite{kerr2023lerf, shafiullah2022clip} (especially in models that take in low-resolution images as input), enabling applications such as $3$D scene-editing \cite{kobayashi2022decomposing, wang2022clip} and robotic manipulation \cite{rashid2023language, shen2023distilled}. Although these methods demonstrate compelling object segmentation performance from open-vocabulary queries, these methods fail to disambiguate between the semantic classes of unique objects that are semantically-similar to the natural language query \cite{kerr2023lerf}. For example, when a user queries specifically for ``tea," existing semantic Gaussian Splatting (GSplat) methods localize the coffee machine without providing any clarifying information on the identity of the localized object, suggesting that ``tea" exists in the scene, when the contrary is true. This ambiguity generally arises from the fact that many vision-language foundation methods, e.g., CLIP, utilize a bag-of-words approach in measuring the semantic similarity between image-language queries. Moreover, many semantic methods for Gaussian Splatting are notably slow, requiring a significant amount of memory and computation time for training and inference. We seek to derive a semantic GSplat method that addresses these limitations.

In this work, we propose \algname, a \emph{Fast, Ambiguity-free Semantics Transfer} method for Gaussian Splatting. \algname generates fine-grained semantic localization, with highly specific semantic classes even in the presence of ambiguous natural-language prompts, and as its name implies, runs much faster than existing semantic GSplat methods. \algname utilizes the vision-language foundation model CLIP \cite{radford2021learning} to embed open-set semantics into GSplat models, enabling optional initialization of relevant objects from closed-set object detectors. However, noting the limited coverage of closed-set object detectors, \algname leverages open-set object detectors and segmentation models to enable queryable semantics from free-form natural language, illustrated in \Cref{fig:architecture}. By constructing a hash-table of identified objects, \algname enables effective semantic disambiguation, while addressing the weaknesses of closed-set methods. To enable significantly faster training and rendering speeds compared to prior work, \algname implements two key innovations: (i) single-phase semantic distillation and (ii) neural-free semantic radiance fields. \algname distills semantics into the underlying GSplat model for scene reconstruction using a single-phase training procedure, simultaneously training the geometric, visual, and semantic components of the GSplat model. In contrast, prior work, e.g., \cite{qin2024langsplat}, utilizes a two-phase training procedure, training the geometric and visual components of the scene before introducing and optimizing the semantic parameters, negatively impacting computational performance. Moreover, unlike 
existing semantic GSplat methods which rely on neural-field models for semantic distillation, \algname directly augments each Gaussian with specific semantic codes, without any neural model. These key insights not only enable \algname to exploit the explicit scene-representation of Gaussian Splatting for faster training and rendering speeds, but also enable \algname to provide unambiguous semantic object labels in response to open-vocabulary user queries, e.g., in semantic object localization, upon identifying semantically-similar objects using the semantic codes associated with each Gaussian, along with the semantic hash-table.

We compare \algname to existing semantic GSplat methods \cite{qin2024langsplat, zhou2024feature} on standard benchmark datasets. \algname achieves $6$x to $8$x faster training times and $18$x to $51$x faster rendering speeds, while requiring $6$x smaller GPU memory. Meanwhile, \algname achieves competitive or better semantic object localization performance. Notably, \algname enables semantic disambiguation in object localization, even with ambiguous user-provided natural-language queries, which we illustrate in \Cref{fig:teaser}. In contrast to prior work, \algname provides a clarifying semantic label to each localized object, disambiguating the semantic identity of the localized object. For example, in \Cref{fig:teaser}, when prompted with the query ``tea," \algname notifies a user that it localized a \emph{coffee machine}, and an \emph{apple} when given the query: ``fruit." Such disambiguation can be critical to enabling interesting downstream applications, e.g., in robotics.

To summarize our contributions:
\begin{itemize}
    \item We introduce \algname, a Fast, Ambiguity-Free Semantic GSplat method that enables notably faster ($6$x to $8$x faster) language-semantics grounding in $3$D scenes, distilled from $2$D vision models.
    \item \algname resolves semantic ambiguity in semantic object localization, arising from ambiguous user-provided natural-language queries or inherent scene ambiguity, enabling the identification of the precise semantic label of relevant objects.
    \item \algname achieves superior rendering speeds for semantic Gaussian Splatting, about $18$x to $51$x faster rendering speeds compared to prior work.
\end{itemize}

\section{Related Work}
\label{sec:related_work}
We provide a detailed review of existing work on open-vocabulary object detection and segmentation, along with follow-on work on grounding $2$D object detection and segmentation in $3$D by leveraging high-fidelity $3$D scene representations.

\smallskip
\noindent \textbf{Open-Vocabulary Object Detection and Segmentation.}
The introduction of the vision transformer (ViT) architecture \cite{dosovitskiy2020image} has spurred rapid research advances in many classical computer vision tasks, such as object detection and segmentation. However, like their convolutional neural network (CNN) counterparts \cite{ren2016faster, he2017mask}, early ViT models \cite{carion2020end, zhang2022dino} were primarily trained to detect objects within a fixed number of classes, defined as closed-set object detection. Vision-image foundation models, e.g., CLIP \cite{radford2021learning} and transformer-based language encoders, such as BERT \cite{Devlin2019BERTPO}, have enabled the detection of objects through open-vocabulary natural-language prompts, without relying on a fixed number of object classes, referred to as open-set object detection. Open-vocabulary object detection methods \cite{li2022grounded, liu2023grounding, minderer2022simple, zareian2021open, gu2021open} fuse the image and language embeddings from vision or language foundation models using a decoder transformer architecture with self-attention to detect objects in a query image. GLIP \cite{li2022grounded} and OWL-ViT \cite{minderer2022simple} train an image and text encoder jointly to learn useful representations, which are fed to a decoder for $2$D grounding. While GLIP and OWL-ViT require a pretraining phase on large-scale internet image-text pairs, GroundingDINO \cite{liu2023grounding} leverages pre-trained text encoder BERT \cite{Devlin2019BERTPO}, but otherwise utilizes a similar model architecture.

Zero-shot object segmentation methods \cite{wang2022open, liu2022open, li2023mask, ravi2024sam} leverage an encoder-decoder model architecture to generate a semantic mask of all instances of objects in an image. Mask DINO \cite{li2023mask} introduces a mask prediction branch to DINO \cite{zhang2022dino}, extending it to object segmentation. Like GLIP, ViL-Seg \cite{liu2022open} trains an image and text encoder together with a clustering head on image-caption pairs gathered on the internet to segment objects, enabling segmentation from natural-language prompts. SAM-2 \cite{ravi2024sam} utilizes an image encoder with memory attention and a memory bank to enable video segmentation (in addition to image segmentation). However, SAM requires a mask, bounding box, or pixel-coordinate prompt for segmentation. GroundedSAM \cite{ren2024grounded} extends SAM-2 to open-vocabulary queries by computing the bounding box associated with a given natural-language query, which is passed as an input prompt to SAM-2. Other methods \cite{li2022language, liang2023open, zhang2023simple} leverage pre-trained image and text encoders, which are fine-tuned in some cases, to enable open-vocabulary object segmentation.

\smallskip
\noindent \textbf{Radiance Fields.}
Neural Radiance Fields (NeRFs) \cite{mildenhall2021nerf, muller2022instant, barron2021mip} represent a scene as a color and density field, parameterized by multi-layer perceptrons. NeRFs generate a photorealistic reconstruction of a scene, capturing high-fidelity visual details. However, high-quality NeRFs typically require long training times and achieve slow rendering rates. Gaussian Splatting \cite{kerbl20233d} addresses these limitations by representing a scene using ellipsoidal primitives with visual attributes, such as opacity and color using spherical harmonics, in addition to the spatial and geometric attributes, such as the ellipsoid's mean and covariance. Radiance fields can be trained entirely from RGB images, making them suitable for many vision tasks.

\smallskip
\noindent \textbf{Grounding Language in 3D.}
To bridge the gap between $2$D object detection and $3$D object localization, prior work has examined grounding semantic embeddings from pre-trained image and text encoders in the $3$D world. CLIP-Fields \cite{shafiullah2022clip} learns an implicit spatial mapping from $3$D points to CLIP image embeddings and Sentence-Bert \cite{reimers2019sentence} text embeddings using back-projected RGB-D points. VLMaps \cite{huang2023visual} and NLMap \cite{chen2023open} distill vision-language features into grid-based maps. The sparsity of these maps limits the resolution of the open-vocabulary segmentation results achievable by these methods. For high-resolution object segmentation, Distilled Features Fields (DFF) \cite{kobayashi2022decomposing}, CLIP-NeRF \cite{wang2022clip}, and LERF \cite{kerr2023lerf} ground CLIP \cite{radford2021learning}, LSeg \cite{li2022language}, and DINO \cite{zhang2022dino} features in a NeRF representation of the scene. Moreover, similar techniques have been employed in \cite{qin2024langsplat, hu2024semantic, zhou2024feature, zuo2024fmgs} to distill semantic features into GSplat representations.
The resulting distilled radiance fields (DRFs) enable dense open-vocabulary object segmentation in $3$D and have been applied to scene-editing \cite{kobayashi2022decomposing, wang2022clip} and robotic mapping and manipulation \cite{rashid2023language, shen2023distilled, zheng2024gaussiangrasper, shorinwa2024splat, shorinwa2025siren}.
Despite their success in object segmentation, existing DRFs fail to precisely identify the semantic class of the segmented object, e.g., for a given query, these methods localize all objects related to the query without resolving the unique semantic identity of each object.
This ambiguity creates a gap in scene understanding, which is often required in many downstream tasks, e.g., in robotic exploration or manipulation. In this work, we introduce a DRF with fine-grained semantics, enabling precise semantic identification of objects in a scene, disambiguating language-guided semantic object localization, and addressing existing challenges.

\section{{Preliminaries}}
\label{sec:preliminaries}
Here, we provide some background information on Gaussian Splatting.
Gaussian Splatting \cite{kerbl20233d} represents a scene using ellipsoidal primitives, with ellipsoid $i$ parameterized by its mean ${\mu_{i} \in \mbb{R}^{3}}$; covariance ${\Sigma = RSS^{\T}R^{\T} \in \mbb{R}^{3 \times 3}}$, where $R$ denotes a rotation matrix, and $S$ denotes a diagonal scaling matrix; opacity $\alpha$; and spherical harmonics (SH). Gaussian Splatting utilizes a tile-based rasterization procedure for efficient real-time rendering rates, outperforming NeRF-based representations, with the color $C$ of each pixel given by the $\alpha$-based blending procedure:
\begin{equation}
    C = \sum_{i \in \mcal{N}} c_{i} \alpha_{i} \prod_{j = 1}{i - 1} (1 - \alpha_{j}),
\end{equation}
where $c_{i}$ denotes the color of each ellipsoid and $\alpha_{i}$ denotes the per-ellipsoid opacity multiplied by the probability density of the $2$D Gaussian associated with the ellipsoid.
Upon initialization from a sparse point cloud, which can be obtained from structure-from-motion, the attributes of each ellipsoid are optimized through gradient descent on a dataset of RGB images. Distilled feature fields build upon NeRFs and Gaussian Splatting, leveraging the same volumetric rendering and tile-based rasterization for NeRFs and Gaussian Splatting, respectively, to generate $2$D feature maps from the $3$D feature fields.

\begin{figure*}[th]
    \centering
    \includegraphics[width=\linewidth, trim={2.5ex 0 0 0}, clip]{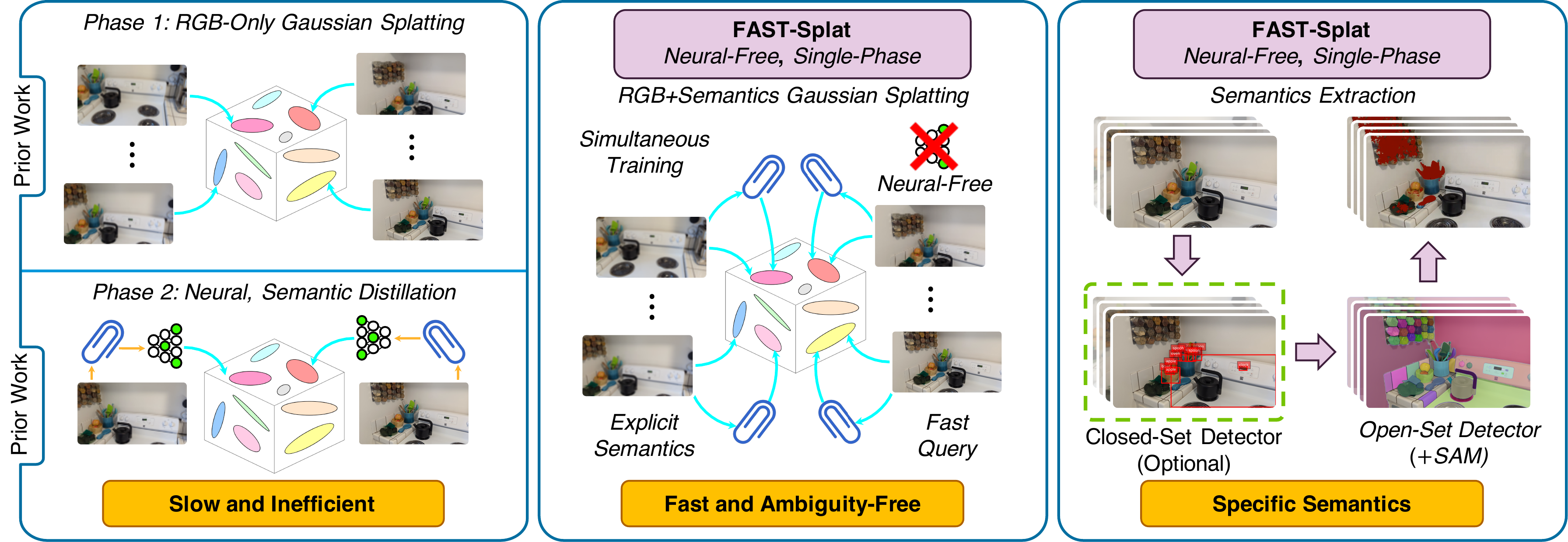}
    \caption{Unlike prior semantic Gaussian Splatting methods, \algname jointly optimizes the geometric, visual, and semantic attributes of Gaussian Splatting models, achieving faster training and rendering times with effective semantics disambiguation using an efficient semantics extraction procedure.}
    \label{fig:architecture}
\end{figure*}

\section{Fast, Ambiguity-Free Semantic Transfer}
\label{sec:method}
Now, we discuss \algname, a fast, fine-grained semantic distillation method for grounding $2$D language semantics in $3$D, enabling precise identification of the semantic label of objects in the $3$D scenes. We build off $3$D Gaussian Splatting to obtain a high-fidelity, photorealistic scene reconstruction entirely from RGB images. 
Given a dataset of images $\mcal{D}$, Gaussian Splatting optimizes the mean, covariance, opacity, and SH parameters of each ellipsoid comprising the representation via the loss function:
\begin{equation}
    \label{eq:gs_loss_function}
    \mcal{L}_{\mathrm{gs}} = (1 - \lambda) \sum_{\mcal{I} \in \mcal{D}} \norm{\mcal{I} - \hat{\mcal{I}}}_{1} + \lambda \mcal{L}_{\mathrm{D-SSIM}},
\end{equation}
where ${\lambda \in [0, 1]}$ denotes a weighting parameter between the ${\ell_{1}}$-loss term and the differentiable structural similarity index measure (D-SSIM) loss term, and $\mcal{I}$ and $\hat{\mcal{I}}$ denote the ground-truth and predicted RGB image, respectively. The GSplat model is trained using the Adam optimizer \cite{Kingma2014AdamAM}, a gradient-based optimization method.

Almost all existing semantic distillation methods rely on neural architectures for grounding language in $3$D in NeRFs \cite{wang2022clip, kobayashi2022decomposing, kerr2023lerf} and GSplat scenes \cite{qin2024langsplat, hu2024semantic, zhou2024feature, zuo2024fmgs}. Although effective in enabling open-set semantic object localization, neural-based approaches introduce significant computational bottlenecks into the training pipeline of NeRF and GSplat scenes. In particular, when utilized in Gaussian Splatting, neural-based semantic distillation essentially erases the computational gains, e.g., faster training times and rendering rates, afforded by Gaussian Splatting when compared to NeRFs. Moreover, prior semantic GSplat methods, e.g., \cite{qin2024langsplat}, train the semantic attributes of a scene independently from its geometric and visual attributes, utilizing a two-phase training procedure, where the geometric and visual attributes are first optimized before the semantic attributes. This training framework significantly increases the time required for training GSplat models. We contrast this training framework with the training framework utilized by \algname, which implements a single-phase training procedure, jointly optimizing all attributes of the scene, in \Cref{fig:architecture}.

Via \algname, we seek to derive a method for open-set semantic distillation, while preserving the faster training times and rendering rates achieved by the underlying GSplat method. 
To achieve this goal, we revisit the fundamental insight exploited by the authors of the original GSplat work \cite{kerbl20233d}, namely: that only representing occupied space explicitly can lead to significant speedup in training and rendering times without any notable degradation in the visual fidelity of the representation.
Specifically, Gaussian Splatting harnesses the explicit form and sparsity of an ellipsoidal-primitive-based scene representation to achieve faster training times and rendering times compared to NeRFs, the state-of-the-art at the time, which relied on implicit fields for the color and density of the scene. 
By learning neural encoders for dimensionality reduction or neural fields for language semantics, existing semantic distillation methods for Gaussian Splatting fail to take advantage of this insight. In contrast, in \algname, we leverage this insight to speed up training and rendering times, by eschewing neural models.
In particular, we augment the attributes of each ellipsoid with a semantic parameter, associated with its semantic identity ${\beta \in \mbb{R}^{3}}$, which is trained simultaneously with the ellipsoid's other attributes. Note that the dimension of $\beta$ does not pose a limitation unique to this work. In fact, existing semantic GSplat methods, e.g., \cite{qin2024langsplat}, generally use three-dimensional semantic codes to enable the utilization of the rasterization procedure employed in Gaussian Splatting.

\smallskip
\noindent \textbf{Closed-Set Semantic Gaussian Splatting.}
\algname supports initialization of the semantics extraction procedure using closed-set object detectors. To obtain the closed-set semantic categories, we feed each image ${I \in \mcal{D}}$ into an object detector, e.g., YOLO \cite{wang2025yolov9}, DETR \cite{carion2020end}, and Efficient DETR \cite{yao2021efficient}, to generate a list of objects present in the image. Although closed-set object detectors generally exhibit remarkable performance, these detectors occasionally fail to identify all instances of an object in an image, given that the size of their dictionary of detectable objects is often limited. 
To overcome this challenge, \algname utilizes an open-set detector, e.g., GroundingDINO \cite{liu2023grounding}, with input prompts from, e.g., image-tagging models \cite{zhang2024recognize} or closed-set object detectors. 
We examine the importance of the open-set object detector in ablations in the experiments in \Cref{sec:evaluation}.

At this stage, each image has a list of constituent objects and bounding boxes specifying the location of these objects in the image. However, high-fidelity semantic distillation generally requires pixel-wise semantic classes. To generate dense semantic maps, we utilize SAM-2 \cite{ravi2024sam}, passing in each object class and its associated bounding box. We maintain a dictionary (hash-table), representing the set of objects identified in the image. Now, we can distill the pixel-wise semantic information into the underlying GSplat model. To predict the semantic class of each ellipsoid in the scene, we apply an affine transformation $\mcal{W}$ to the semantic attribute $\beta$ of each ellipsoid, mapping $\beta$ to an $N$-dimensional space, where $N$ denotes the size of the dictionary. Subsequently, we apply the softmax function to the resulting outputs to define a probability over the entries in the dictionary. We extend this procedure to view-based rendering. Given a camera pose for rendering, we utilize differentiable tile-based rasterization \cite{kerbl20233d} to render a semantic feature map
corresponding to the camera pose. We utilize the aforementioned procedure to transform the semantic feature of each pixel to a probability distribution over the entries in the hash-table.
We optimize the semantic attributes using the multi-class cross-entropy loss function, given by:
\begin{equation}
    \mcal{L}_{\mathrm{ce}} = -\sum_{\mcal{I} \in \mcal{D}} \frac{1}{\vert \mcal{T} \vert} \sum_{p \in \mcal{T}} \gamma_{p} \log \left(\frac{\exp(\mcal{F}_{\mcal{I}, p, c})}{\sum_{j = 1}^{N} \exp(\mcal{F}_{\mcal{I},p, j})} \right),
\end{equation}
where $\mcal{T}$ represents the set of all indices of pixels in the rendered feature map $\mcal{F}_{\mcal{I}}$ associated with image $\mcal{I}$ (after applying the affine transformation $\mcal{W}$), $c$ denotes the true class of the pixel, and $\gamma_{p}$ represents the weight associated with pixel $p$. We augment the set of object classes for each view with an \emph{undetected} class label and assign a lower weight to pixels with this class label, i.e., pixels that were not segmented during the data pre-processing phase. This augmentation step serves as a regularization term for the stability of undetected pixels. We train all attributes of the GSplat representation, e.g., the geometric, visual, and semantic attributes and $\mcal{W}$, simultaneously using the loss function:
${\mcal{L}_{\mathrm{sgs}} = \mcal{L}_{\mathrm{gs}} + 
    \mcal{L}_{\mathrm{ce}},}$
with $\mcal{L}_{\mathrm{gs}}$ defined in \eqref{eq:gs_loss_function}.

\smallskip
\noindent \textbf{Open-Vocabulary Semantic Gaussian Splatting.}
The resulting distillation method 
enables object localization from a pre-defined set of object classes. However, it does not support open-ended queries from users, which represent an important practical use-case. Here, we extend the capability of the distillation method from the closed-set setting to the open-vocabulary setting, leveraging open-source pre-trained text encoders, e.g., CLIP \cite{radford2021learning} and BERT \cite{devlin2018bert}. These pre-trained text encoders compute text embeddings for natural-language inputs, mapping these inputs to a metric space. 

We pre-compute the text embeddings $\theta_{\mathrm{d}}$ of each entry in the dictionary of detected object classes using the pre-trained text encoder, without any fine-tuning. At runtime, given a natural-language prompt specified by a user, \algname computes the text embedding $\theta_{\mathrm{q}}$ associated with the prompt using the pre-trained text encoder. 
To identify semantically-relevant objects in the scene, we compute the \emph{relevancy score} \cite{kerr2023lerf}, which represents the pairwise softmax between the cosine-similarity values computed over the dictionary embeddings $\theta_{\mathrm{d}}$ and a set of text embeddings consisting of the query embedding $\theta_{\mathrm{q}}$ (\emph{positive} embedding) and a set of canonical embeddings (\emph{negative} embeddings).

Given a threshold on the relevancy score, we designate all entries whose relevancy score exceeds this specified threshold as being similar to the query. We map these entries to pixels in a rendered feature map using the probability distribution over all entries computed via the softmax function. In general, we can assign a label to each pixel by taking the maximum probability value; however, alternative techniques can be utilized, including multi-hypothesis prediction methods, e.g., conformal prediction \cite{shafer2008tutorial}.

\smallskip
\noindent \textbf{Fine-grained Language Semantics.}
In the real-world, identifying only the location of a related object of interest, e.g., in semantic object localization, is generally inadequate. Identifying the precise semantic class of the localized object is of even greater importance, especially for downstream applications where semantic object localization constitutes a single component of a solution pipeline. Prior semantic GSplat methods lack this critical capability. For example, existing semantic GSplat methods will localize a coffee machine when a user asks specifically for some ``tea,"  without disambiguating the semantic identity of the localized object, which could be misleading to a user. \algname seeks to address this limitation.
The novel distillation technique employed by \algname endows the algorithm with the capability to not only identify the precise semantic class of each object, but also to disambiguate the semantic identity of related objects. 
Given a natural-language prompt, \algname identifies the related object classes from its dictionary and subsequently, provides the location of these objects in the scene along with their specific semantic label, illustrated in \Cref{fig:teaser}.

\section{Experiments}
\label{sec:evaluation}
We examine the semantic segmentation and disambiguation capabilities of \algname, e.g., in scenarios with ambiguous user prompts and scenarios with multiple related objects. We summarize the results here and provide additional implementation details and results in the Supplementary Material.

\noindent\textbf{Baselines.} We evaluate \algname against existing semantic GSplat methods: LangSplat \cite{qin2024langsplat} and Feature-3DGS (\mbox{F-3DGS}) \cite{zhou2024feature}. We do not compare our work with FMGS \cite{zuo2024fmgs}, which does not have a publicly-available code implementation. We utilize the original code implementation provided by the authors of these baseline methods.

\noindent\textbf{Datasets and Metrics.}  We evaluate each method across the benchmark datasets: 3D-OVS \cite{liu2023weakly}, LERF \cite{kerr2023lerf}, and \mbox{MipNeRF360} \cite{barron2022mip}. Further, we examine each method in uncurated real-world scenes, e.g., in typical household settings. In each dataset, we report the mean intersection-over-union (mIoU), localization accuracy, and training time [mins].

\subsection{Comparison: Benchmark Datasets.}

\begin{figure*}[th]
    \centering
    \includegraphics[width=\linewidth]{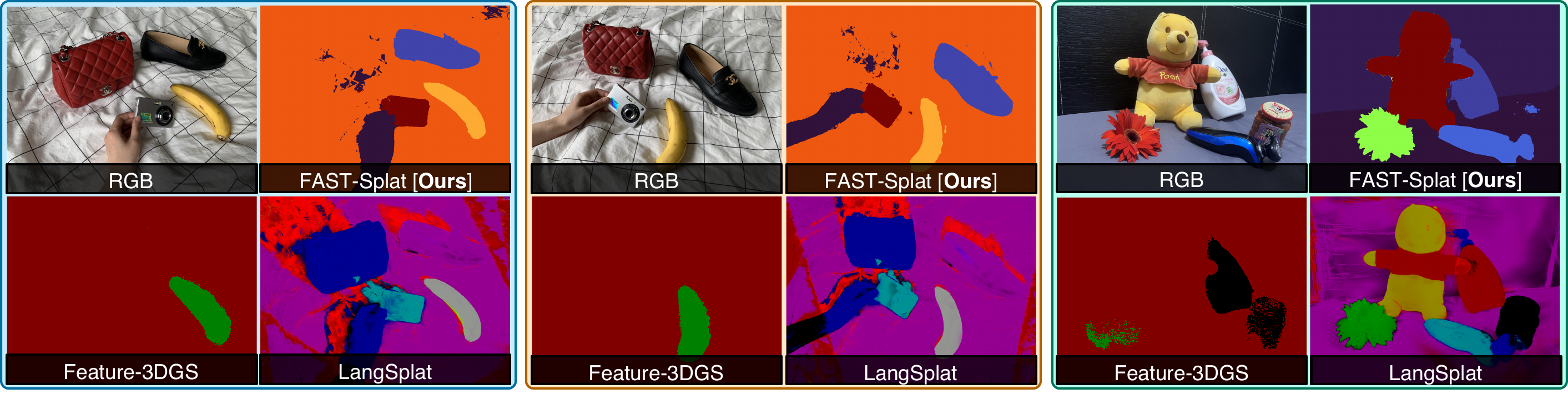}
    \caption{Semantic segmentation on the \emph{Bed} and \emph{Covered-Desk} scenes in 3D-OVS. Overall, \algname outperforms the baselines.}
    \label{fig:3dovs_experiments_seg_bed_covered_desk}
\end{figure*}

We evaluate all methods on the 3D-OVS dataset, composed of a set of long-tail objects in different backgrounds and poses, and the LERF dataset.
\Cref{tab:seg_perf_miou_3dovs} presents the mIoU scores achieved by each method in the 3D-OVS dataset, with \algname achieving the highest mIoU in each scene. Specifically, \algname outperforms LangSplat and F-3DGS by about $8.3\%$ and $40\%$, respectively, in mIoU. Similarly, from \Cref{tab:seg_perf_acc_3dovs}, \algname achieves the highest localization accuracy scores, outperforming LangSplat and F-3DGS by about $1.2\%$ and $7.4\%$, respectively. 

Moreover, the novel architecture of \algname enables significantly faster training, highlighted in \Cref{tab:train_time_perf_3dovs}, where \algname achieves $5.6$x and $7.7$x speedups in training time compared to LangSplat and F-3DGS, respectively. LangSplat requires two independent training phases for each scene (one for the underlying Gaussian representation and another for the distillation procedure), resulting in significantly longer training times. In addition, LangSplat utilizes costly data pre-processing procedures, making multiple queries to the CLIP and SAM models for semantics extraction, further degrading its computational performance. Similarly, \mbox{F-3DGS} suffers from these challenges. In contrast, \algname overcomes these challenges with its single-phase training procedure (see \Cref{fig:architecture}), resulting in much faster training times.
These results underscore the superior performance of \algname compared to existing methods. 
In \Cref{fig:3dovs_experiments_seg_bed_covered_desk}, we show the semantic segmentation masks generated by each method. Unlike F-3DGS, which fails to segment many of the objects in the scenes, \algname and LangSplat provide quite accurate segmentation masks. However, on average, \algname achieves higher accuracy.
We present the results on the LERF dataset in Appendix~B.

\begin{table}[tbph]
	\centering
	\caption{mIoU Scores in the 3D-OVS Dataset.}
	\label{tab:seg_perf_miou_3dovs}
	\begin{adjustbox}{width=\linewidth}
		{\begin{tabular}{l c c c}
                \toprule
                Scene & LangSplat \cite{qin2024langsplat} & F-3DGS \cite{zhou2024feature} & FAST-Splat [Ours] \\
                \midrule
                Bed & 0.925 & 0.643 & \textbf{0.933} \\
                Covered-Desk & 0.779 & 0.678 & \textbf{0.913} \\
                \midrule
                overall & 0.852 & 0.661 & \textbf{0.923} \\
                \bottomrule
		\end{tabular}}
	\end{adjustbox}
\end{table}

\begin{table}[tbph]
	\centering
	\caption{Localization accuracy in the 3D-OVS Dataset.}
	\label{tab:seg_perf_acc_3dovs}
	\begin{adjustbox}{width=\linewidth}
		{\begin{tabular}{l c c c}
                \toprule
                Scene & LangSplat \cite{qin2024langsplat} & F-3DGS \cite{zhou2024feature} & FAST-Splat [Ours] \\
                \midrule
                Bed & 0.992 & 0.961 & \textbf{0.998} \\
                Covered-Desk & 0.978 & 0.895 & \textbf{0.995} \\
                \midrule
                overall & 0.985 & 0.928 & \textbf{0.997} \\
                \bottomrule
		\end{tabular}}
	\end{adjustbox}
\end{table}

\begin{table}[tbph]
	\centering
	\caption{Total training times [mins] in the 3D-OVS Dataset.}
	\label{tab:train_time_perf_3dovs}
	\begin{adjustbox}{width=\linewidth}
		{\begin{tabular}{l c c c}
                \toprule
                Scene & LangSplat \cite{qin2024langsplat} & F-3DGS \cite{zhou2024feature} & FAST-Splat [Ours] \\
                \midrule
                Bed & 79.23 & 107.25 & \textbf{13.96} \\
                Covered-Desk & 78.36 & 112.06 & \textbf{13.96} \\
                \bottomrule
		\end{tabular}}
	\end{adjustbox}
\end{table}

\subsection{Ablations.}
Next, we ablate the choice of using an open-set detector compared to using only a closed-set detector in \algname using the MipNerf360 datasets. We use the YOLO model for closed-set object detection.  In \Cref{tab:ablation_seg_perf_miou_mipnerf360,tab:ablation_seg_perf_acc_mipnerf360}, we report the mIoU and localization accuracy associated with each approach. From these results, the open-set detector significantly improves the segmentation performance of \algname beyond that of the closed-set detector. In particular, the open-set detector improves the overall mIoU and localization accuracy scores by about $97\%$ and $25\%$, respectively, compared to the closed-set detector. This performance boost can be explained by the greater expressiveness of the open-set detector, enabling it to identify varied objects, compared to the relatively limited set of objects that can be identified by a closed-set detector. However, using an open-set detector results in slightly longer training times. The closed-set approach requires about $10.30$~mins and $25.73$~mins in the \emph{Kitchen} and \emph{Bicycle} scenes, respectively, compared to the open-set approach, which requires about $12.28$~mins and $27.82$~mins in the \emph{Kitchen} and \emph{Bicycle} scenes.
We provide further ablation results, along with renderings of the segmentation masks, using the 3D-OVS and LERF datasets in Appendix~C.

\begin{table}[tbph]
	\centering
        \begin{minipage}{.48\columnwidth}
            \centering
    	\caption{IoU Scores in \mbox{MipNeRF360}.}
    	\label{tab:ablation_seg_perf_miou_mipnerf360}
    	\begin{adjustbox}{width=\linewidth}
    		{
                \begin{tabular}{l c c}
                    \toprule
                    Scene & Closed-Set & Open-Set \\
                    \midrule
                    Kitchen & 0.366 & \textbf{0.869} \\
                    Bicycle & 0.395 & \textbf{0.633} \\
                    \midrule
                    overall & 0.381 & \textbf{0.751} \\
                    \bottomrule
    		\end{tabular}
            }
    	\end{adjustbox}
        \end{minipage}%
        \hfill
        \begin{minipage}{.48\columnwidth}
            \centering
    	\caption{Localization \mbox{accuracy} in MipNeRF360.}
    	\label{tab:ablation_seg_perf_acc_mipnerf360}
    	\begin{adjustbox}{width=\linewidth}
    		{
                \begin{tabular}{l c c}
                    \toprule
                    Scene & Closed-Set & Open-Set \\
                    \midrule
                    Kitchen & 0.887 & \textbf{0.976} \\
                    Bicycle & 0.577 & \textbf{0.855} \\
                    \midrule
                    overall & 0.732 & \textbf{0.916} \\
                    \bottomrule
    		\end{tabular}
            }
    	\end{adjustbox}
        \end{minipage}
\end{table}

\begin{figure*}[th]
    \centering
    \includegraphics[width=\linewidth]{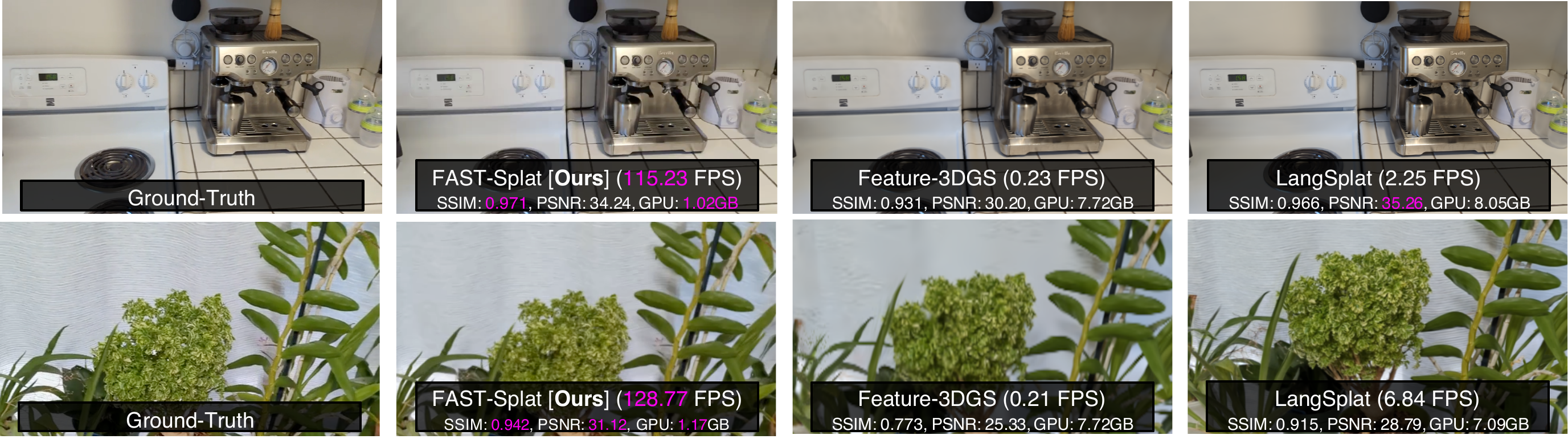}
    \caption{RGB images rendered in \emph{R-Kitchen} and \emph{R-Library}. \algname achieves $18$x to $51$x faster rendering speeds with at least $6$x lower memory usage, while achieving competitive or better reconstruction quality.}
    \label{fig:experiments_rgb}
\end{figure*}

\subsection{Comparison: Uncurated Real-World Datasets.}
In addition, we create a dataset of images from uncurated real-world scenes to capture the diversty of objects that would typically occur in real-world scene. We collect data from (i) a home kitchen in its natural, everyday state, comprising the \emph{R-Kitchen} dataset and (ii) a home library, containing books and plants, comprising the \emph{R-Library} dataset. We keep the scale of the training images in the uncurated dataset the same as that of MipNerf360, a widely-used benchmark dataset. We describe the data setup in greater detail in the Appendix~A.
Here, we examine the fidelity of the $3$D scene reconstruction, reporting the structural similarity index measure (SSIM), the peak signal-to-noise ratio (PSNR), the rendering speed in frames per second [FPS], and memory usage [GB] in \Cref{tab:image_quality_metrics}. \algname enables real-time rendering speeds, essentially retaining the performance of the original GSplat work \cite{kerbl20233d}, compared to LangSplat and F-3DGS, whose performance is limited by their semantics distillation design. \algname achieves $51$x and $18.8$x speedup in rendering speed and the highest SSIM scores compared to the best-competing method, in the \emph{R-Kitchen} and \emph{R-Library} datasets, respectively. LangSplat achieves the highest PSNR score in the \emph{R-Kitchen} dataset, while \algname achieves the highest PSNR score in the \emph{R-Library} dataset. Further, \algname reduces the memory usage by about $7.6$x and $6.1$x and the training times by about $6.7$x and $8.4$x, compared to the best competitor in both datasets.

\begin{table}[tbph]
	\centering
	\caption{The rendering frame-rate (Speed [FPS]), image quality, and memory usage (Memory  [GB]) in Uncurated Real-World Datasets.}
	\label{tab:image_quality_metrics}
	\begin{adjustbox}{width=\linewidth}
		{\begin{tabular}{l c c c c}
                \toprule
    
                Method & Speed $\uparrow$ & SSIM $\uparrow$ & PSNR $\uparrow$ & Memory $\downarrow$ \\
                \midrule
                \multicolumn{5}{c}{\emph{R-Kitchen}} \\
                \midrule
                LangSplat \cite{qin2024langsplat} & 2.25 & 0.966 & \textbf{35.26} & 8.05 \\
                F-3DGS \cite{zhou2024feature} & 0.23  & 0.931 & 30.20 &  7.72 \\
                FAST-Splat [\textbf{ours}] & \textbf{115.23} & \textbf{0.971} & 34.24 & \textbf{1.02} \\
                \midrule
                \multicolumn{5}{c}{\emph{R-Library}} \\
                \midrule
                LangSplat \cite{qin2024langsplat} &  6.84 & 0.915 & 28.79 & 7.09 \\
                F-3DGS \cite{zhou2024feature} & 0.21  & 0.773 & 25.33 & 7.72 \\
                FAST-Splat [\textbf{ours}]  & \textbf{128.77} & \textbf{0.942} & \textbf{31.12} & \textbf{1.17} \\
                \bottomrule    
		\end{tabular}}
	\end{adjustbox}
\end{table}

\begin{table}[tbph]
	\centering
	\caption{Total training times [mins] in Uncurated Real-World Datasets.}
	\label{tab:train_time_perf_unc_rw}
	\begin{adjustbox}{width=\linewidth}
		{\begin{tabular}{l c c c}
                \toprule
                Scene & LangSplat \cite{qin2024langsplat} & F-3DGS \cite{zhou2024feature} & FAST-Splat [Ours] \\
                \midrule
                R-Kitchen & 187.03 & 88.24 & \textbf{13.20} \\
                R-Library & 199.60 & 126.91 & \textbf{15.02} \\
                \bottomrule
		\end{tabular}}
	\end{adjustbox}
\end{table}

\begin{figure*}[th]
    \centering
    \includegraphics[width=\linewidth]{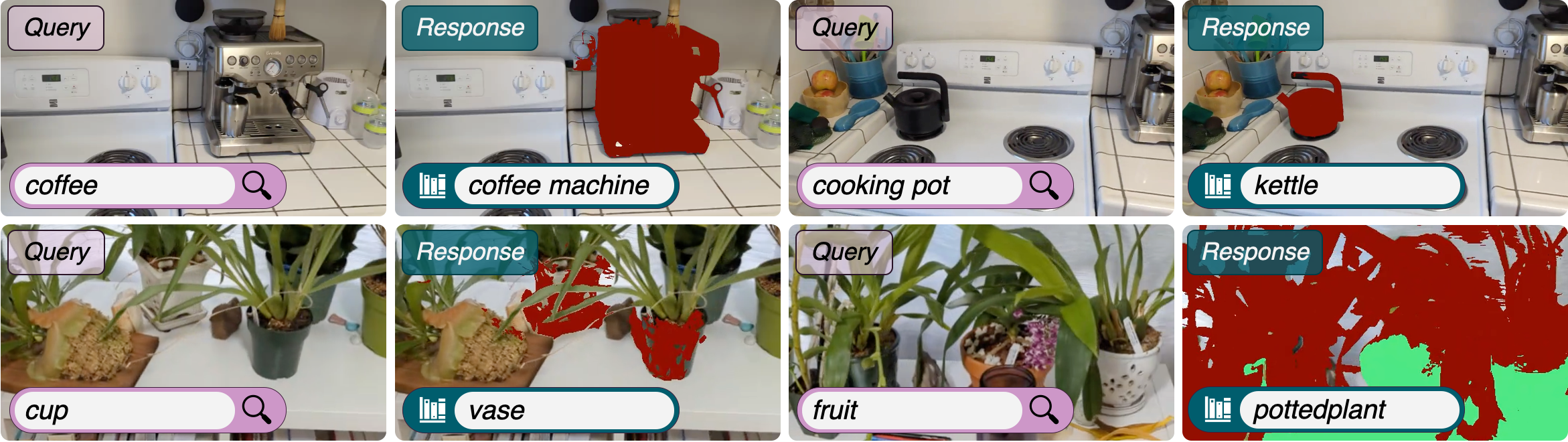}
    \caption{\algname resolves language ambiguity in natural-language queries in semantic object localization, identifying the specific semantic class of each object, e.g., a \emph{coffee machine} and a \emph{kettle}, when prompted with an ambiguous query, e.g., ``coffee" and ``cooking pot," respectively. Likewise, \algname disambiguates between a ``cup" and a \emph{vase} and between a ``fruit" and a \emph{pottedplant} in \emph{R-Library}.}
    \label{fig:experiments_semantic_disambiguation}
\end{figure*}

\smallskip
\noindent \textbf{Semantic Disambiguation.}
\algname extends the capabilities of semantic GSplat methods beyond semantic object localization to the domain of semantic disambiguation. As noted in the preceding discussion, existing semantic GSplat methods cannot resolve semantic ambiguity either arising from the natural-language query or inherently from the scene composition. In contrast, when provided with an ambiguous natural-language query, \algname disambiguates the semantic object localization task, by identifying the specific semantic label of related objects in the scene. 

We provide some examples in \Cref{fig:experiments_semantic_disambiguation}. For example, when a user queries for ``coffee" in the \emph{R-Kitchen} scene, \algname provides the user with the semantic identity of the related object, \emph{coffee machine,} along with its semantic segmentation mask and location, given a threshold for the segmentation task. Likewise, when queried for a ``cooking pot," \algname identifies the semantic class: \emph{kettle}, including its segmentation mask and location. Notably, in the \emph{R-Library} scene, \algname resolves the ambiguity between a cup and a vase. When queried with the prompt ``cup," \algname informs the user that, more precisely, a \emph{vase} exists in the scene. Similarly, when asked to localize a ``fruit," \algname clarifies that the scene contains a \emph{pottedplant} and provides the semantic segmentation mask and location.

\begin{figure}[th]
    \centering
    \includegraphics[width=\linewidth]{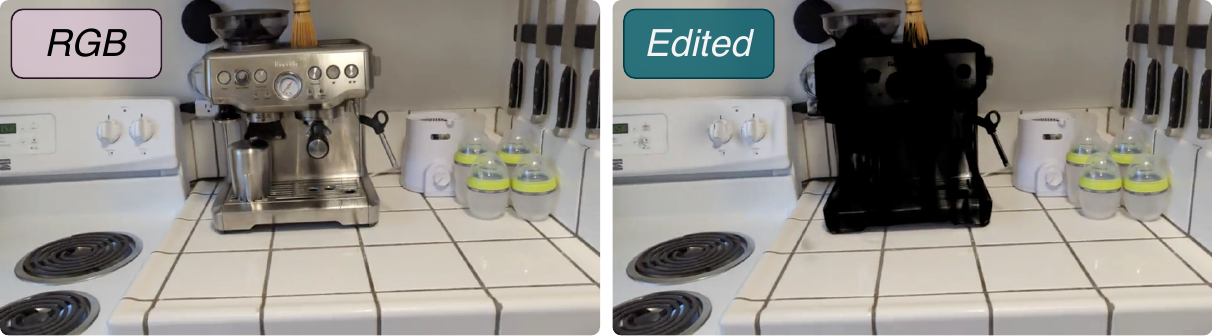}
    \caption{Color-editing of a coffee machine using \algname.}
    \label{fig:scene_editing}
\end{figure}

\smallskip
\noindent \textbf{Scene Editing.}
\algname enables scene-editing of GSplat scenes, by leveraging semantics in Gaussian Splatting. With \algname, a user can modify the visual properties of a scene, and in addition, insert, delete, or move objects in the scene. For space considerations, we omit a detailed discussion of this capability, given that it has been discussed in prior work. In \Cref{fig:scene_editing}, we provide an example where the color of the \emph{coffee machine} in \emph{R-Kitchen} is modified.

\section{Conclusion}
\label{sec:conclusion}
We introduce \algname, a fast semantic GSplat method that resolves ambiguity in semantic object localization. \algname retains the explicit form of Gaussian Splatting and avoids neural architectures to achieve fast training and rendering speeds compared to existing methods. \algname jointly trains the geometric, visual, and semantic attributes of a scene without any neural models, resulting in faster training times, rendering rates. \algname  leverages a hash-table of semantic codes to resolve semantic ambiguity arising from the natural-language query or the scene composition, in the semantic object localization task, addressing one of the fundamental limitations of prior semantic GSplat methods.

\section{Limitations and Future Work}
\label{sec:limitations_future_work}
\algname relies on 2D vision-language foundation models, e.g., CLIP, for distilling semantics into GSplat scenes. Hence, the quality of the resulting semantic GSplat depends on the expressiveness of the semantic embeddings computed by the vision-language model. \algname might fail to recognize objects if the vision-language model fails to correctly identify such objects. Future work will seek to estimate the uncertainty of the semantic embeddings computed by vision-language models to better characterize the limitations of the trained semantic GSplat and ultimately provide safety and performance guarantees for the semantic GSplat.

\section*{Acknowledgment}
This work was supported in part by NSF project FRR 2342246, and ONR project N00014-23-1-2354, and DARPA project HR001120C0107. Toyota Research Institute provided funds to support this work.

{
    \small
    \bibliographystyle{style/ieeenat_fullname}
    \bibliography{references.bib}

\begin{thebibliography}{52}
\providecommand{\natexlab}[1]{#1}
\providecommand{\url}[1]{\texttt{#1}}
\expandafter\ifx\csname urlstyle\endcsname\relax
  \providecommand{\doi}[1]{doi: #1}\else
  \providecommand{\doi}{doi: \begingroup \urlstyle{rm}\Url}\fi

\bibitem[Barron et~al.(2021)Barron, Mildenhall, Tancik, Hedman, Martin-Brualla, and Srinivasan]{barron2021mip}
Jonathan~T Barron, Ben Mildenhall, Matthew Tancik, Peter Hedman, Ricardo Martin-Brualla, and Pratul~P Srinivasan.
\newblock Mip-nerf: A multiscale representation for anti-aliasing neural radiance fields.
\newblock In \emph{Proceedings of the IEEE/CVF international conference on computer vision}, pages 5855--5864, 2021.

\bibitem[Barron et~al.(2022)Barron, Mildenhall, Verbin, Srinivasan, and Hedman]{barron2022mip}
Jonathan~T Barron, Ben Mildenhall, Dor Verbin, Pratul~P Srinivasan, and Peter Hedman.
\newblock Mip-nerf 360: Unbounded anti-aliased neural radiance fields.
\newblock In \emph{Proceedings of the IEEE/CVF conference on computer vision and pattern recognition}, pages 5470--5479, 2022.

\bibitem[Carion et~al.(2020)Carion, Massa, Synnaeve, Usunier, Kirillov, and Zagoruyko]{carion2020end}
Nicolas Carion, Francisco Massa, Gabriel Synnaeve, Nicolas Usunier, Alexander Kirillov, and Sergey Zagoruyko.
\newblock End-to-end object detection with transformers.
\newblock In \emph{European conference on computer vision}, pages 213--229. Springer, 2020.

\bibitem[Chen et~al.(2023)Chen, Xia, Ichter, Rao, Gopalakrishnan, Ryoo, Stone, and Kappler]{chen2023open}
Boyuan Chen, Fei Xia, Brian Ichter, Kanishka Rao, Keerthana Gopalakrishnan, Michael~S Ryoo, Austin Stone, and Daniel Kappler.
\newblock Open-vocabulary queryable scene representations for real world planning.
\newblock In \emph{2023 IEEE International Conference on Robotics and Automation (ICRA)}, pages 11509--11522. IEEE, 2023.

\bibitem[Devlin(2018)]{devlin2018bert}
Jacob Devlin.
\newblock Bert: Pre-training of deep bidirectional transformers for language understanding.
\newblock \emph{arXiv preprint arXiv:1810.04805}, 2018.

\bibitem[Devlin et~al.(2019)Devlin, Chang, Lee, and Toutanova]{Devlin2019BERTPO}
Jacob Devlin, Ming-Wei Chang, Kenton Lee, and Kristina Toutanova.
\newblock Bert: Pre-training of deep bidirectional transformers for language understanding.
\newblock In \emph{North American Chapter of the Association for Computational Linguistics}, 2019.

\bibitem[Dosovitskiy(2020)]{dosovitskiy2020image}
Alexey Dosovitskiy.
\newblock An image is worth 16x16 words: Transformers for image recognition at scale.
\newblock \emph{arXiv preprint arXiv:2010.11929}, 2020.

\bibitem[Gu et~al.(2021)Gu, Lin, Kuo, and Cui]{gu2021open}
Xiuye Gu, Tsung-Yi Lin, Weicheng Kuo, and Yin Cui.
\newblock Open-vocabulary object detection via vision and language knowledge distillation.
\newblock \emph{arXiv preprint arXiv:2104.13921}, 2021.

\bibitem[He et~al.(2017)He, Gkioxari, Doll{\'a}r, and Girshick]{he2017mask}
Kaiming He, Georgia Gkioxari, Piotr Doll{\'a}r, and Ross Girshick.
\newblock Mask r-cnn.
\newblock In \emph{Proceedings of the IEEE international conference on computer vision}, pages 2961--2969, 2017.

\bibitem[Hu et~al.(2024)Hu, Wang, Fan, Fan, Peng, Lei, Li, and Zhang]{hu2024semantic}
Xu Hu, Yuxi Wang, Lue Fan, Junsong Fan, Junran Peng, Zhen Lei, Qing Li, and Zhaoxiang Zhang.
\newblock Semantic anything in 3d gaussians.
\newblock \emph{arXiv preprint arXiv:2401.17857}, 2024.

\bibitem[Huang et~al.(2023)Huang, Mees, Zeng, and Burgard]{huang2023visual}
Chenguang Huang, Oier Mees, Andy Zeng, and Wolfram Burgard.
\newblock Visual language maps for robot navigation.
\newblock In \emph{2023 IEEE International Conference on Robotics and Automation (ICRA)}, pages 10608--10615. IEEE, 2023.

\bibitem[Kerbl et~al.(2023)Kerbl, Kopanas, Leimk{\"u}hler, and Drettakis]{kerbl20233d}
Bernhard Kerbl, Georgios Kopanas, Thomas Leimk{\"u}hler, and George Drettakis.
\newblock 3d gaussian splatting for real-time radiance field rendering.
\newblock \emph{ACM Trans. Graph.}, 42\penalty0 (4):\penalty0 139--1, 2023.

\bibitem[Kerr et~al.(2023)Kerr, Kim, Goldberg, Kanazawa, and Tancik]{kerr2023lerf}
Justin Kerr, Chung~Min Kim, Ken Goldberg, Angjoo Kanazawa, and Matthew Tancik.
\newblock Lerf: Language embedded radiance fields.
\newblock In \emph{Proceedings of the IEEE/CVF International Conference on Computer Vision}, pages 19729--19739, 2023.

\bibitem[Kingma and Ba(2014)]{Kingma2014AdamAM}
Diederik~P. Kingma and Jimmy Ba.
\newblock Adam: A method for stochastic optimization.
\newblock \emph{CoRR}, abs/1412.6980, 2014.

\bibitem[Kobayashi et~al.(2022)Kobayashi, Matsumoto, and Sitzmann]{kobayashi2022decomposing}
Sosuke Kobayashi, Eiichi Matsumoto, and Vincent Sitzmann.
\newblock Decomposing nerf for editing via feature field distillation.
\newblock \emph{Advances in Neural Information Processing Systems}, 35:\penalty0 23311--23330, 2022.

\bibitem[Li et~al.(2022{\natexlab{a}})Li, Weinberger, Belongie, Koltun, and Ranftl]{li2022language}
Boyi Li, Kilian~Q Weinberger, Serge Belongie, Vladlen Koltun, and Ren{\'e} Ranftl.
\newblock Language-driven semantic segmentation.
\newblock \emph{arXiv preprint arXiv:2201.03546}, 2022{\natexlab{a}}.

\bibitem[Li et~al.(2023)Li, Zhang, Xu, Liu, Zhang, Ni, and Shum]{li2023mask}
Feng Li, Hao Zhang, Huaizhe Xu, Shilong Liu, Lei Zhang, Lionel~M Ni, and Heung-Yeung Shum.
\newblock Mask dino: Towards a unified transformer-based framework for object detection and segmentation.
\newblock In \emph{Proceedings of the IEEE/CVF Conference on Computer Vision and Pattern Recognition}, pages 3041--3050, 2023.

\bibitem[Li et~al.(2022{\natexlab{b}})Li, Zhang, Zhang, Yang, Li, Zhong, Wang, Yuan, Zhang, Hwang, et~al.]{li2022grounded}
Liunian~Harold Li, Pengchuan Zhang, Haotian Zhang, Jianwei Yang, Chunyuan Li, Yiwu Zhong, Lijuan Wang, Lu Yuan, Lei Zhang, Jenq-Neng Hwang, et~al.
\newblock Grounded language-image pre-training.
\newblock In \emph{Proceedings of the IEEE/CVF Conference on Computer Vision and Pattern Recognition}, pages 10965--10975, 2022{\natexlab{b}}.

\bibitem[Liang et~al.(2023)Liang, Wu, Dai, Li, Zhao, Zhang, Zhang, Vajda, and Marculescu]{liang2023open}
Feng Liang, Bichen Wu, Xiaoliang Dai, Kunpeng Li, Yinan Zhao, Hang Zhang, Peizhao Zhang, Peter Vajda, and Diana Marculescu.
\newblock Open-vocabulary semantic segmentation with mask-adapted clip.
\newblock In \emph{Proceedings of the IEEE/CVF Conference on Computer Vision and Pattern Recognition}, pages 7061--7070, 2023.

\bibitem[Liu et~al.(2023{\natexlab{a}})Liu, Zhan, Zhang, Xu, Yu, El~Saddik, Theobalt, Xing, and Lu]{liu2023weakly}
Kunhao Liu, Fangneng Zhan, Jiahui Zhang, Muyu Xu, Yingchen Yu, Abdulmotaleb El~Saddik, Christian Theobalt, Eric Xing, and Shijian Lu.
\newblock Weakly supervised 3d open-vocabulary segmentation.
\newblock \emph{Advances in Neural Information Processing Systems}, 36:\penalty0 53433--53456, 2023{\natexlab{a}}.

\bibitem[Liu et~al.(2022)Liu, Wen, Han, Xu, Xu, and Liang]{liu2022open}
Quande Liu, Youpeng Wen, Jianhua Han, Chunjing Xu, Hang Xu, and Xiaodan Liang.
\newblock Open-world semantic segmentation via contrasting and clustering vision-language embedding.
\newblock In \emph{European Conference on Computer Vision}, pages 275--292. Springer, 2022.

\bibitem[Liu et~al.(2023{\natexlab{b}})Liu, Zeng, Ren, Li, Zhang, Yang, Li, Yang, Su, Zhu, et~al.]{liu2023grounding}
Shilong Liu, Zhaoyang Zeng, Tianhe Ren, Feng Li, Hao Zhang, Jie Yang, Chunyuan Li, Jianwei Yang, Hang Su, Jun Zhu, et~al.
\newblock Grounding dino: Marrying dino with grounded pre-training for open-set object detection.
\newblock \emph{arXiv preprint arXiv:2303.05499}, 2023{\natexlab{b}}.

\bibitem[Mildenhall et~al.(2021)Mildenhall, Srinivasan, Tancik, Barron, Ramamoorthi, and Ng]{mildenhall2021nerf}
Ben Mildenhall, Pratul~P Srinivasan, Matthew Tancik, Jonathan~T Barron, Ravi Ramamoorthi, and Ren Ng.
\newblock Nerf: Representing scenes as neural radiance fields for view synthesis.
\newblock \emph{Communications of the ACM}, 65\penalty0 (1):\penalty0 99--106, 2021.

\bibitem[Minderer et~al.(2022)Minderer, Gritsenko, Stone, Neumann, Weissenborn, Dosovitskiy, Mahendran, Arnab, Dehghani, Shen, et~al.]{minderer2022simple}
Matthias Minderer, Alexey Gritsenko, Austin Stone, Maxim Neumann, Dirk Weissenborn, Alexey Dosovitskiy, Aravindh Mahendran, Anurag Arnab, Mostafa Dehghani, Zhuoran Shen, et~al.
\newblock Simple open-vocabulary object detection.
\newblock In \emph{European Conference on Computer Vision}, pages 728--755. Springer, 2022.

\bibitem[M{\"u}ller et~al.(2022)M{\"u}ller, Evans, Schied, and Keller]{muller2022instant}
Thomas M{\"u}ller, Alex Evans, Christoph Schied, and Alexander Keller.
\newblock Instant neural graphics primitives with a multiresolution hash encoding.
\newblock \emph{ACM transactions on graphics (TOG)}, 41\penalty0 (4):\penalty0 1--15, 2022.

\bibitem[Qin et~al.(2024)Qin, Li, Zhou, Wang, and Pfister]{qin2024langsplat}
Minghan Qin, Wanhua Li, Jiawei Zhou, Haoqian Wang, and Hanspeter Pfister.
\newblock Langsplat: 3d language gaussian splatting.
\newblock In \emph{Proceedings of the IEEE/CVF Conference on Computer Vision and Pattern Recognition}, pages 20051--20060, 2024.

\bibitem[Radford et~al.(2021)Radford, Kim, Hallacy, Ramesh, Goh, Agarwal, Sastry, Askell, Mishkin, Clark, et~al.]{radford2021learning}
Alec Radford, Jong~Wook Kim, Chris Hallacy, Aditya Ramesh, Gabriel Goh, Sandhini Agarwal, Girish Sastry, Amanda Askell, Pamela Mishkin, Jack Clark, et~al.
\newblock Learning transferable visual models from natural language supervision.
\newblock In \emph{International conference on machine learning}, pages 8748--8763. PMLR, 2021.

\bibitem[Rashid et~al.(2023)Rashid, Sharma, Kim, Kerr, Chen, Kanazawa, and Goldberg]{rashid2023language}
Adam Rashid, Satvik Sharma, Chung~Min Kim, Justin Kerr, Lawrence~Yunliang Chen, Angjoo Kanazawa, and Ken Goldberg.
\newblock Language embedded radiance fields for zero-shot task-oriented grasping.
\newblock In \emph{7th Annual Conference on Robot Learning}, 2023.

\bibitem[Ravi et~al.(2024)Ravi, Gabeur, Hu, Hu, Ryali, Ma, Khedr, R{\"a}dle, Rolland, Gustafson, et~al.]{ravi2024sam}
Nikhila Ravi, Valentin Gabeur, Yuan-Ting Hu, Ronghang Hu, Chaitanya Ryali, Tengyu Ma, Haitham Khedr, Roman R{\"a}dle, Chloe Rolland, Laura Gustafson, et~al.
\newblock Sam 2: Segment anything in images and videos.
\newblock \emph{arXiv preprint arXiv:2408.00714}, 2024.

\bibitem[Reimers(2019)]{reimers2019sentence}
N Reimers.
\newblock Sentence-bert: Sentence embeddings using siamese bert-networks.
\newblock \emph{arXiv preprint arXiv:1908.10084}, 2019.

\bibitem[Ren et~al.(2016)Ren, He, Girshick, and Sun]{ren2016faster}
Shaoqing Ren, Kaiming He, Ross Girshick, and Jian Sun.
\newblock Faster r-cnn: Towards real-time object detection with region proposal networks.
\newblock \emph{IEEE transactions on pattern analysis and machine intelligence}, 39\penalty0 (6):\penalty0 1137--1149, 2016.

\bibitem[Ren et~al.(2024)Ren, Liu, Zeng, Lin, Li, Cao, Chen, Huang, Chen, Yan, et~al.]{ren2024grounded}
Tianhe Ren, Shilong Liu, Ailing Zeng, Jing Lin, Kunchang Li, He Cao, Jiayu Chen, Xinyu Huang, Yukang Chen, Feng Yan, et~al.
\newblock Grounded sam: Assembling open-world models for diverse visual tasks.
\newblock \emph{arXiv preprint arXiv:2401.14159}, 2024.

\bibitem[Sch\"{o}nberger and Frahm(2016)]{schoenberger2016sfm}
Johannes~Lutz Sch\"{o}nberger and Jan-Michael Frahm.
\newblock Structure-from-motion revisited.
\newblock In \emph{Conference on Computer Vision and Pattern Recognition (CVPR)}, 2016.

\bibitem[Shafer and Vovk(2008)]{shafer2008tutorial}
Glenn Shafer and Vladimir Vovk.
\newblock A tutorial on conformal prediction.
\newblock \emph{Journal of Machine Learning Research}, 9\penalty0 (3), 2008.

\bibitem[Shafiullah et~al.(2022)Shafiullah, Paxton, Pinto, Chintala, and Szlam]{shafiullah2022clip}
Nur Muhammad~Mahi Shafiullah, Chris Paxton, Lerrel Pinto, Soumith Chintala, and Arthur Szlam.
\newblock Clip-fields: Weakly supervised semantic fields for robotic memory.
\newblock \emph{arXiv preprint arXiv:2210.05663}, 2022.

\bibitem[Shen et~al.(2023)Shen, Yang, Yu, Wong, Kaelbling, and Isola]{shen2023distilled}
William Shen, Ge Yang, Alan Yu, Jansen Wong, Leslie~Pack Kaelbling, and Phillip Isola.
\newblock Distilled feature fields enable few-shot language-guided manipulation.
\newblock \emph{arXiv preprint arXiv:2308.07931}, 2023.

\bibitem[Shorinwa et~al.(2024)Shorinwa, Tucker, Smith, Swann, Chen, Firoozi, Kennedy~III, and Schwager]{shorinwa2024splat}
Ola Shorinwa, Johnathan Tucker, Aliyah Smith, Aiden Swann, Timothy Chen, Roya Firoozi, Monroe Kennedy~III, and Mac Schwager.
\newblock Splat-mover: Multi-stage, open-vocabulary robotic manipulation via editable gaussian splatting.
\newblock \emph{arXiv preprint arXiv:2405.04378}, 2024.

\bibitem[Shorinwa et~al.(2025)Shorinwa, Sun, Schwager, and Majumdar]{shorinwa2025siren}
Ola Shorinwa, Jiankai Sun, Mac Schwager, and Anirudha Majumdar.
\newblock Siren: Semantic, initialization-free registration of multi-robot gaussian splatting maps.
\newblock \emph{arXiv preprint arXiv:2502.06519}, 2025.

\bibitem[Tancik et~al.(2023)Tancik, Weber, Ng, Li, Yi, Wang, Kristoffersen, Austin, Salahi, Ahuja, et~al.]{tancik2023nerfstudio}
Matthew Tancik, Ethan Weber, Evonne Ng, Ruilong Li, Brent Yi, Terrance Wang, Alexander Kristoffersen, Jake Austin, Kamyar Salahi, Abhik Ahuja, et~al.
\newblock Nerfstudio: A modular framework for neural radiance field development.
\newblock In \emph{ACM SIGGRAPH 2023 Conference Proceedings}, pages 1--12, 2023.

\bibitem[Wang et~al.(2022{\natexlab{a}})Wang, Chai, He, Chen, and Liao]{wang2022clip}
Can Wang, Menglei Chai, Mingming He, Dongdong Chen, and Jing Liao.
\newblock Clip-nerf: Text-and-image driven manipulation of neural radiance fields.
\newblock In \emph{Proceedings of the IEEE/CVF Conference on Computer Vision and Pattern Recognition}, pages 3835--3844, 2022{\natexlab{a}}.

\bibitem[Wang et~al.(2025)Wang, Yeh, and Mark~Liao]{wang2025yolov9}
Chien-Yao Wang, I-Hau Yeh, and Hong-Yuan Mark~Liao.
\newblock Yolov9: Learning what you want to learn using programmable gradient information.
\newblock In \emph{European Conference on Computer Vision}, pages 1--21. Springer, 2025.

\bibitem[Wang et~al.(2022{\natexlab{b}})Wang, Feiszli, Wang, Malik, and Tran]{wang2022open}
Weiyao Wang, Matt Feiszli, Heng Wang, Jitendra Malik, and Du Tran.
\newblock Open-world instance segmentation: Exploiting pseudo ground truth from learned pairwise affinity.
\newblock In \emph{Proceedings of the IEEE/CVF conference on computer vision and pattern recognition}, pages 4422--4432, 2022{\natexlab{b}}.

\bibitem[Yao et~al.(2021)Yao, Ai, Li, and Zhang]{yao2021efficient}
Zhuyu Yao, Jiangbo Ai, Boxun Li, and Chi Zhang.
\newblock Efficient detr: improving end-to-end object detector with dense prior.
\newblock \emph{arXiv preprint arXiv:2104.01318}, 2021.

\bibitem[Yu et~al.(2022)Yu, Wang, Vasudevan, Yeung, Seyedhosseini, and Wu]{yu2022coca}
Jiahui Yu, Zirui Wang, Vijay Vasudevan, Legg Yeung, Mojtaba Seyedhosseini, and Yonghui Wu.
\newblock Coca: Contrastive captioners are image-text foundation models.
\newblock \emph{arXiv preprint arXiv:2205.01917}, 2022.

\bibitem[Zareian et~al.(2021)Zareian, Rosa, Hu, and Chang]{zareian2021open}
Alireza Zareian, Kevin~Dela Rosa, Derek~Hao Hu, and Shih-Fu Chang.
\newblock Open-vocabulary object detection using captions.
\newblock In \emph{Proceedings of the IEEE/CVF Conference on Computer Vision and Pattern Recognition}, pages 14393--14402, 2021.

\bibitem[Zhang et~al.(2022)Zhang, Li, Liu, Zhang, Su, Zhu, Ni, and Shum]{zhang2022dino}
Hao Zhang, Feng Li, Shilong Liu, Lei Zhang, Hang Su, Jun Zhu, Lionel~M Ni, and Heung-Yeung Shum.
\newblock Dino: Detr with improved denoising anchor boxes for end-to-end object detection.
\newblock \emph{arXiv preprint arXiv:2203.03605}, 2022.

\bibitem[Zhang et~al.(2023)Zhang, Li, Zou, Liu, Li, Yang, and Zhang]{zhang2023simple}
Hao Zhang, Feng Li, Xueyan Zou, Shilong Liu, Chunyuan Li, Jianwei Yang, and Lei Zhang.
\newblock A simple framework for open-vocabulary segmentation and detection.
\newblock In \emph{Proceedings of the IEEE/CVF International Conference on Computer Vision}, pages 1020--1031, 2023.

\bibitem[Zhang et~al.(2024)Zhang, Huang, Ma, Li, Luo, Xie, Qin, Luo, Li, Liu, et~al.]{zhang2024recognize}
Youcai Zhang, Xinyu Huang, Jinyu Ma, Zhaoyang Li, Zhaochuan Luo, Yanchun Xie, Yuzhuo Qin, Tong Luo, Yaqian Li, Shilong Liu, et~al.
\newblock Recognize anything: A strong image tagging model.
\newblock In \emph{Proceedings of the IEEE/CVF Conference on Computer Vision and Pattern Recognition}, pages 1724--1732, 2024.

\bibitem[Zheng et~al.(2024)Zheng, Chen, Zheng, Gu, Yang, Jin, Li, Zhong, Wang, Liu, et~al.]{zheng2024gaussiangrasper}
Yuhang Zheng, Xiangyu Chen, Yupeng Zheng, Songen Gu, Runyi Yang, Bu Jin, Pengfei Li, Chengliang Zhong, Zengmao Wang, Lina Liu, et~al.
\newblock Gaussiangrasper: 3d language gaussian splatting for open-vocabulary robotic grasping.
\newblock \emph{arXiv preprint arXiv:2403.09637}, 2024.

\bibitem[Zhou et~al.(2024)Zhou, Chang, Jiang, Fan, Zhu, Xu, Chari, You, Wang, and Kadambi]{zhou2024feature}
Shijie Zhou, Haoran Chang, Sicheng Jiang, Zhiwen Fan, Zehao Zhu, Dejia Xu, Pradyumna Chari, Suya You, Zhangyang Wang, and Achuta Kadambi.
\newblock Feature 3dgs: Supercharging 3d gaussian splatting to enable distilled feature fields.
\newblock In \emph{Proceedings of the IEEE/CVF Conference on Computer Vision and Pattern Recognition}, pages 21676--21685, 2024.

\bibitem[Zhou et~al.(2022)Zhou, Girdhar, Joulin, Kr{\"a}henb{\"u}hl, and Misra]{zhou2022detecting}
Xingyi Zhou, Rohit Girdhar, Armand Joulin, Philipp Kr{\"a}henb{\"u}hl, and Ishan Misra.
\newblock Detecting twenty-thousand classes using image-level supervision.
\newblock In \emph{European Conference on Computer Vision}, pages 350--368. Springer, 2022.

\bibitem[Zuo et~al.(2024)Zuo, Samangouei, Zhou, Di, and Li]{zuo2024fmgs}
Xingxing Zuo, Pouya Samangouei, Yunwen Zhou, Yan Di, and Mingyang Li.
\newblock Fmgs: Foundation model embedded 3d gaussian splatting for holistic 3d scene understanding.
\newblock \emph{International Journal of Computer Vision}, pages 1--17, 2024.

\end{thebibliography}
}

\clearpage

\appendix
\begin{appendices}
    \section{Implementation Details}
\label{sec:appendix_experiment_implementation}
To implement \algname, we build off Splatfacto in Nerfstudio \cite{tancik2023nerfstudio}, and utilize the text encoder of the OpenAI's CLIP ResNet model \emph{RN50x64} \cite{radford2021learning} to compute the text embeddings. However, other text encoders can also be used, e.g., other CLIP models or BERT \cite{devlin2018bert}. We train \algname using Stochastic Gradient Descent with momentum, via the Adam optimizer \cite{Kingma2014AdamAM}.

For the uncurated real-world datasets, we collected the training data by taking videos of the scene using a smartphone camera. We utilized structure-from-motion, e.g., COLMAP \cite{schoenberger2016sfm}, for computing the pose of the camera for each frame used in training the GSplat model.
\algname is trained on an NVIDIA GeForce RTX 3090 GPU with 24GB VRAM for $30000$ iterations, using the hyperparameters specified in \cite{tancik2023nerfstudio} in its Splafacto model.
We train the baseline methods LangSplat and Feature-3DGS
using the original code implementation provided by their respective authors on an NVIDIA GeForce RTX 3090 Ti GPU with 24GB VRAM, giving LangSplat and Feature-3DGS an advantage on hardware, since the 3090 Ti has $10752$ CUDA cores compared to the $10496$ CUDA cores on the 3090, among other improved hardware specifications.
To train LangSplat, we follow the procedure provided by its authors, which involves pre-training the Gaussian Splatting Scene using the code implementation provided in \cite{kerbl20233d}, prior to training the semantic components. We do not include the structure-from-motion processing times in any of the results reported in the experiments, given that this procedure is required universally by all methods.

\section{Comparison on the LERF Dataset}
We provide additional results on the segmentation performance of each method in the LERF dataset. In \Cref{tab:seg_perf_miou_lerf,tab:seg_perf_acc_lerf,tab:train_time_perf_lerf}, we provide the mIoU and localization accuracy scores achieved by \algname, LangSplat, and \mbox{F-3DGS}. \algname achieves the highest overall mIoU score and localization accuracy and achieves the highest localization accuracy in each scene. F-3DGS achieves the highest mIoU score in the \emph{Ramen} and \emph{Waldo\_kitchen} scenes, while \algname achieves the highest mIoU scores in the \emph{Teatime} and \emph{Waldo\_kitchen} scenes. From \Cref{tab:train_time_perf_lerf}, \algname requires significantly shorter training times compared to the baselines, providing $7.7$x, $6.9$x, and $7.5$x speedups compared to the best-competing methods in the \emph{Ramen}, \emph{Teatime}, and \emph{Waldo\_kitchen} scenes, respectively. These results further highlight the superior performance of \algname compared to existing baselines.

\begin{table}[tbph]
	\centering
	\caption{mIoU Scores in the LERF Dataset.}
	\label{tab:seg_perf_miou_lerf}
	\begin{adjustbox}{width=\linewidth}
		{\begin{tabular}{l c c c}
                \toprule
                Scene & LangSplat \cite{qin2024langsplat} & F-3DGS \cite{zhou2024feature} & FAST-Splat [Ours] \\
                \midrule
                Ramen & 0.512 & \textbf{0.600} & 0.566 \\
                Teatime & 0.651 & 0.613 & \textbf{0.706} \\
                Waldo\_kitchen & 0.445 & \textbf{0.533} & \textbf{0.533} \\
                \midrule
                overall & 0.536 & 0.582 & \textbf{0.602} \\
                \bottomrule
		\end{tabular}}
	\end{adjustbox}
\end{table}

\begin{table}[tbph]
	\centering
	\caption{Localization accuracy in the LERF Dataset.}
	\label{tab:seg_perf_acc_lerf}
	\begin{adjustbox}{width=\linewidth}
		{\begin{tabular}{l c c c}
                \toprule
                Scene & LangSplat \cite{qin2024langsplat} & F-3DGS \cite{zhou2024feature} & FAST-Splat [Ours] \\
                \midrule
                Ramen & 0.732 & 0.876 & \textbf{0.935} \\
                Teatime & 0.881 & 0.890 & \textbf{0.972} \\
                Waldo\_kitchen & 0.955 & 0.863 & \textbf{0.971} \\
                \midrule
                overall & 0.856 & 0.876 & \textbf{0.959} \\
                \bottomrule
		\end{tabular}}
	\end{adjustbox}
\end{table}

\begin{table}[tbph]
	\centering
	\caption{Total training times [mins] in the LERF Dataset.}
	\label{tab:train_time_perf_lerf}
	\begin{adjustbox}{width=\linewidth}
		{\begin{tabular}{l c c c}
                \toprule
                Scene & LangSplat \cite{qin2024langsplat} & F-3DGS \cite{zhou2024feature} & FAST-Splat [Ours] \\
                \midrule
                Ramen & 141.98 & 116.49 & \textbf{15.08} \\
                Teatime & 175.26 & 114.53 & \textbf{16.70} \\
                Waldo\_kitchen & 189.55 & 123.77 & \textbf{16.59} \\
                \bottomrule
		\end{tabular}}
	\end{adjustbox}
\end{table}

\begin{figure*}[th]
    \centering
    \begin{minipage}{.48\textwidth}
        \centering
        \includegraphics[width=\linewidth]{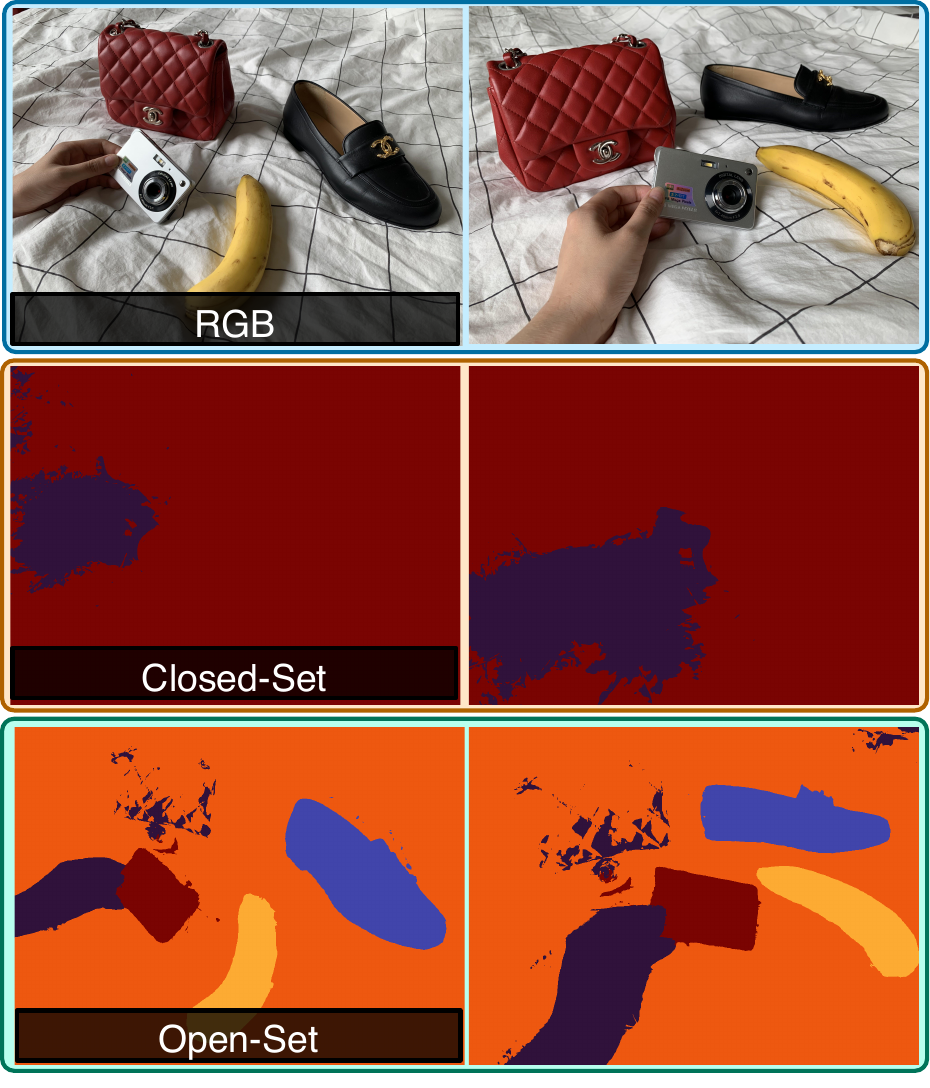}
        \caption{Semantic segmentation on the \emph{Bed} scene in 3D-OVS. Whereas the closed-set approach fails to produce high-quality segmentation masks, the open-set approach achieves high accuracy.}
        \label{fig:experiments_seg_3dovs_bed}
    \end{minipage}%
    \hfill
    \begin{minipage}{.48\textwidth}
        \centering
        \includegraphics[width=\linewidth]{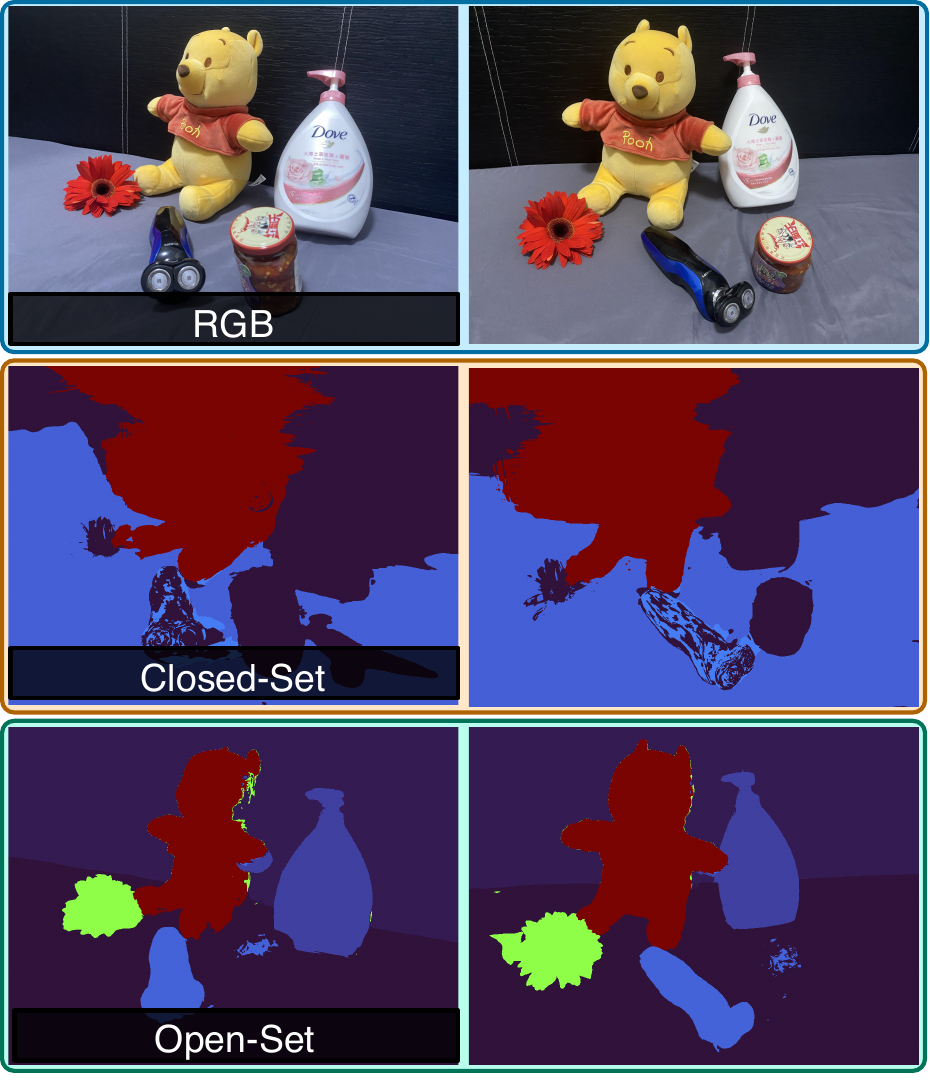}
        \caption{Semantic segmentation on the \emph{Covered-Desk} in 3D-OVS. The open-set approach achieves higher accuracy, especially for long-tail objects, e.g., the \emph{gerbera} (a daisy/flower).}
        \label{fig:experiments_seg_3dovs_covered_desk}
    \end{minipage}
\end{figure*}

\section{Ablations.}
Further, we discuss ablation results in the 3D-OVS and LERF datasets, where we evaluate the performance of \algname with only a closed-set detector and the variant with an open-set detector.
\Cref{tab:ablation_seg_perf_miou_3dovs,tab:ablation_seg_perf_acc_3dovs} provide the mIoU and localization accuracy scores for the closed-set and open-set approaches in the 3D-OVS dataset. In this dataset, the open-set approach improves the mIoU score by about $8.8$x and the localization accuracy by about $1.9$x relative to that of the closed-set approach.
In \Cref{fig:experiments_seg_3dovs_bed,fig:experiments_seg_3dovs_covered_desk}, we show the segmentation masks produced by each approach. The closed-set approach fails to produce meaningful segmentation masks in the \emph{Bed} scene; in contrast, the open-set approach produces highly-accurate segmentation masks. The closed-set approach performs better in the \emph{Covered-Desk} scene (relative to the \emph{Bed} scene), localizing the teddy bear. However, the closed-set approach fails when segmenting long-tail objects such as the \mbox{\emph{gerbera}} in \Cref{fig:experiments_seg_3dovs_covered_desk}, which the open-set approach successfully localizes.

\begin{figure*}[th]
    \centering
    \includegraphics[width=\linewidth]{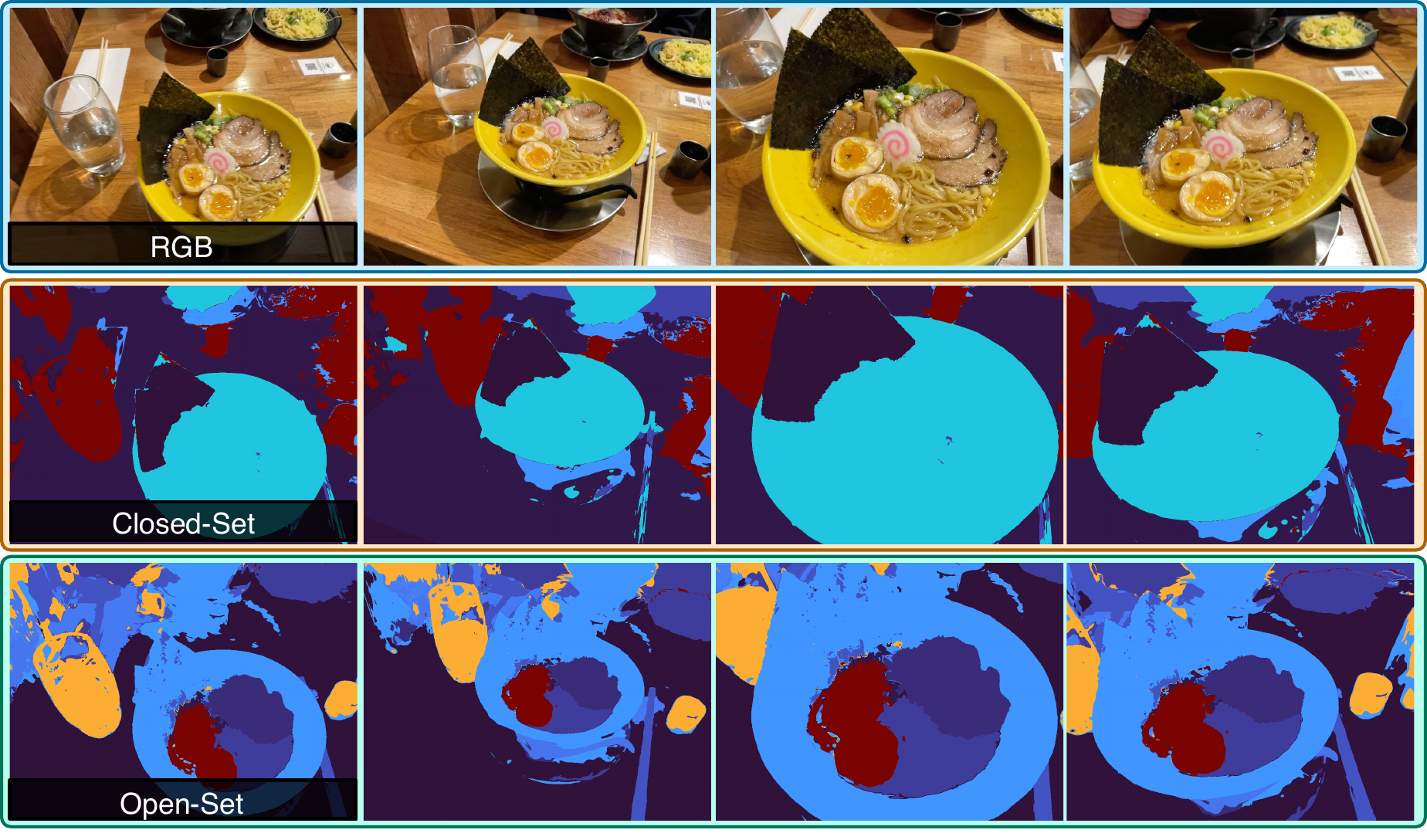}
    \caption{Semantic segmentation of the \emph{Ramen} scene in the LERF dataset. The closed-set approach fails to generate highly-detailed segmentation masks, unlike the open-set approach.}
    \label{fig:experiments_seg_lerf_ramen}
\end{figure*}

\begin{figure*}[th]
    \centering
    \includegraphics[width=\linewidth]{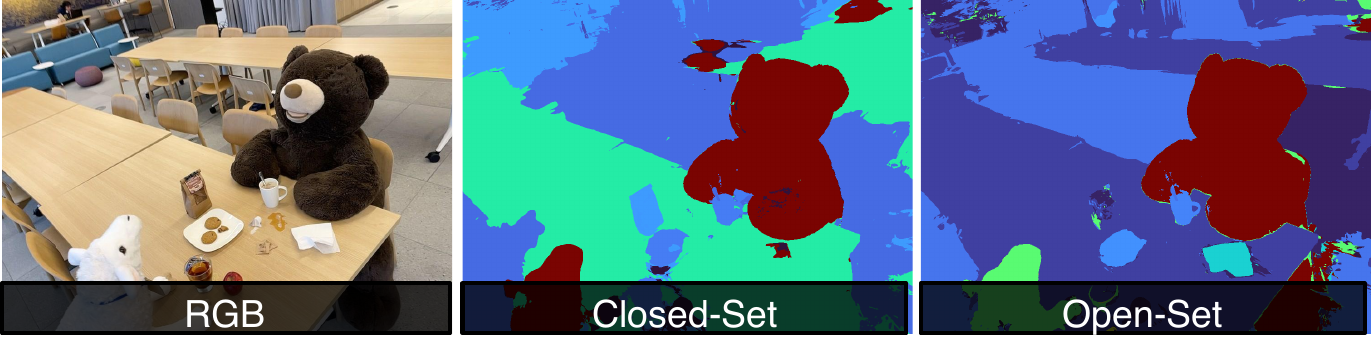}
    \caption{Semantic segmentation of the \emph{Teatime} scene in the LERF dataset. The closed-set achieves a relatively better segmentation performance compared to other scenes, but fails to distinguish between the sheep and the teddy bear.}
    \label{fig:experiments_seg_lerf_teatime}
\end{figure*}

We observe similar findings in the LERF dataset in \Cref{tab:ablation_seg_perf_miou_lerf,tab:ablation_seg_perf_acc_lerf}, where the open-set approach improves the mIoU and localization accuracy scores by about $2$x and $1.15$x.
\Cref{fig:experiments_seg_lerf_ramen,fig:experiments_seg_lerf_teatime} shows the segmentation masks generated by each approach. While the closed-set approach fails to produce highly-detailed segmentation masks, e.g., segmenting the egg, noodles, and pork-belly in the bowl, the open-set approach provides higher-fidelity segmentation masks. Likewise, the closed-set approach fails to distinguish the sheep from the teddy bear in \Cref{fig:experiments_seg_lerf_teatime}, unlike the open-set approach.
However, in all the datasets, the closed-set approach requires shorter training times compared to the open-set approach.

\begin{figure*}[th]
    \centering
    \includegraphics[width=\linewidth]{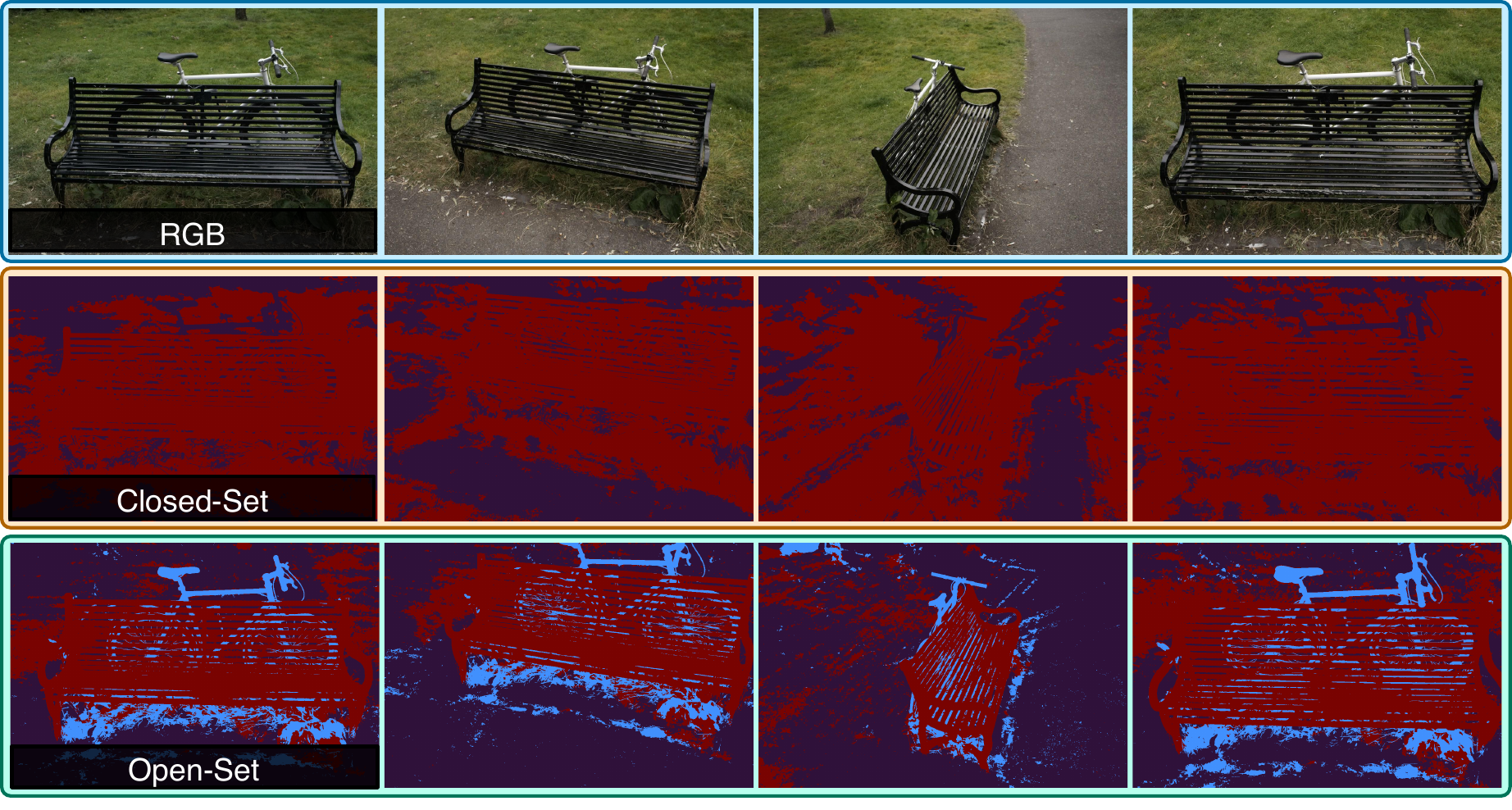}
    \caption{Semantic segmentation of the \emph{Bicycle} scene in the MipNerf360 dataset. The closed-set approach fails to accurately segment the bicycle behind the bench, unlike the open-set approach.}
    \label{fig:experiments_seg_mipnerf360_bicycle}
\end{figure*}

Lastly, in \Cref{fig:experiments_seg_mipnerf360_bicycle}, we provide a few segmentation images generated by the closed-set and open-set approaches in the MipNerf360 datasets. Although the closed-set approach fails to accurately segment the bicycle behind the bench, the open-set approach successfully segments the bicycle.
These results underscore the importance of the open-set detector in \algname.

\begin{table}[tbph]
	\centering
	\caption{mIoU Scores in the 3D-OVS Datasets.}
	\label{tab:ablation_seg_perf_miou_3dovs}
	\begin{adjustbox}{width=0.65\linewidth}
		{\begin{tabular}{l c c}
                \toprule
                Scene & Closed-Set & Open-Set \\
                \midrule
                Bed & 0.044 & \textbf{0.933} \\
                Covered-Desk & 0.165 & \textbf{0.913} \\
                \midrule
                overall & 0.105 & \textbf{0.923} \\
                \bottomrule
		\end{tabular}}
	\end{adjustbox}
\end{table}

\begin{table}[tbph]
	\centering
	\caption{Localization accuracy in the 3D-OVS Datasets.}
	\label{tab:ablation_seg_perf_acc_3dovs}
	\begin{adjustbox}{width=0.65\linewidth}
		{\begin{tabular}{l c c}
                \toprule
                Scene & Closed-Set & Open-Set \\
                \midrule
                Bed & 0.165 & \textbf{0.998} \\
                Covered-Desk & 0.893 & \textbf{0.995} \\
                \midrule
                overall & 0.529 & \textbf{0.997} \\
                \bottomrule
		\end{tabular}}
	\end{adjustbox}
\end{table}

\begin{table}[tbph]
	\centering
	\caption{Total training times [mins] in the 3D-OVS Datasets.}
	\label{tab:ablation_train_time_perf_3dovs}
	\begin{adjustbox}{width=0.65\linewidth}
		{\begin{tabular}{l c c}
                \toprule
                Scene & Closed-Set & Open-Set \\
                \midrule
                Bed & \textbf{12.6} & 13.96 \\
                Covered-Desk & \textbf{12.07} & 13.96 \\
                \bottomrule
		\end{tabular}}
	\end{adjustbox}
\end{table}

\begin{table}[tbph]
	\centering
	\caption{mIoU Scores in the LERF Datasets.}
	\label{tab:ablation_seg_perf_miou_lerf}
	\begin{adjustbox}{width=0.65\linewidth}
		{\begin{tabular}{l c c}
                \toprule
                Scene & Closed-Set & Open-Set \\
                \midrule
                Ramen & 0.135 & 0.566 \\
                Teatime & 0.407 & \textbf{0.706} \\
                Waldo\_kitchen & 0.343 & \textbf{0.533} \\
                \midrule
                overall & 0.295 & \textbf{0.602} \\
                \bottomrule
		\end{tabular}}
	\end{adjustbox}
\end{table}

\begin{table}[tbph]
	\centering
	\caption{Localization accuracy in the LERF Datasets.}
	\label{tab:ablation_seg_perf_acc_lerf}
	\begin{adjustbox}{width=0.65\linewidth}
		{\begin{tabular}{l c c}
                \toprule
                Scene & Closed-Set & Open-Set \\
                \midrule
                Ramen & 0.736 & \textbf{0.935} \\
                Teatime & 0.916 & \textbf{0.972} \\
                Waldo\_kitchen & 0.858 & \textbf{0.971} \\
                \midrule
                overall & 0.837 & \textbf{0.959} \\
                \bottomrule
		\end{tabular}}
	\end{adjustbox}
\end{table}

\begin{table}[tbph]
	\centering
	\caption{Total training times [mins] in the LERF Datasets.}
	\label{tab:ablation_train_time_perf_lerf}
	\begin{adjustbox}{width=0.65\linewidth}
		{\begin{tabular}{l c c}
                \toprule
                Scene & Closed-Set & Open-Set \\
                \midrule
                Ramen & \textbf{13.86} & 15.08 \\
                Teatime & \textbf{15.01} & 16.70 \\
                Waldo\_kitchen & \textbf{13.99} & 16.59 \\
                \bottomrule
		\end{tabular}}
	\end{adjustbox}
\end{table}

\section{Semantic Object Localization}
\label{sec:appendix_experiment_semantic_localization}
\begin{figure*}[th]
    \centering
    \includegraphics[width=\linewidth]{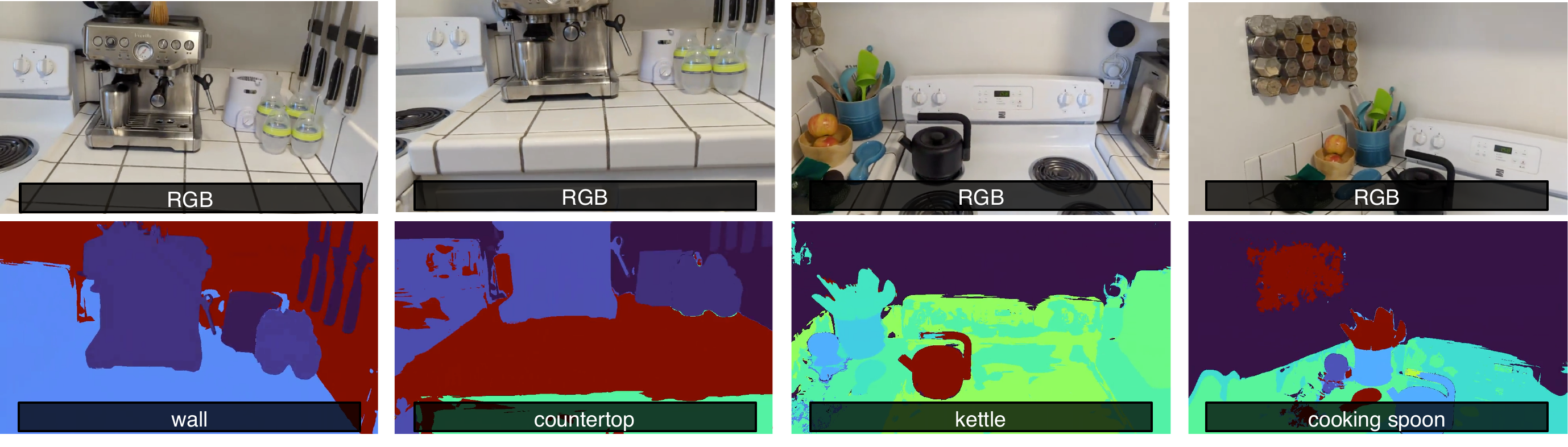}
    \caption{\algname enables semantic object localization wth open-vocabulary user prompts. Here, we provide segmentation results for a ``wall," ``countertop," ``kettle," and ``cooking spoon."}
    \label{fig:app_experiments_localization}
\end{figure*}

Here, we provide additional results, demonstrating the photorealistic novel-view synthesis capability of \algname, including its semantic segmentation masks. In \Cref{fig:app_experiments_localization}, we show the semantic localization of a ``wall," ``countertop," ``kettle," and ``cooking spoon." In each case, \algname successfully localizes the pertinent object. The colors in each figure vary with the similarity of the object to the natural-language query. We utilize the ``Turbo" colormap, with an increase in the intensity of the red color channel indicating an \emph{increase} in the similarity of an object to the query and an increase in the intensity of the blue color channel indicating a \emph{decrease} in the relevance of an object to the query, in general. Moreover, the RGB images rendered by \algname appear photorealistic, displayed in the top row in \Cref{fig:app_experiments_localization}, all rendered at novel views.

\begin{figure*}[th]
    \centering
    \includegraphics[width=\linewidth]{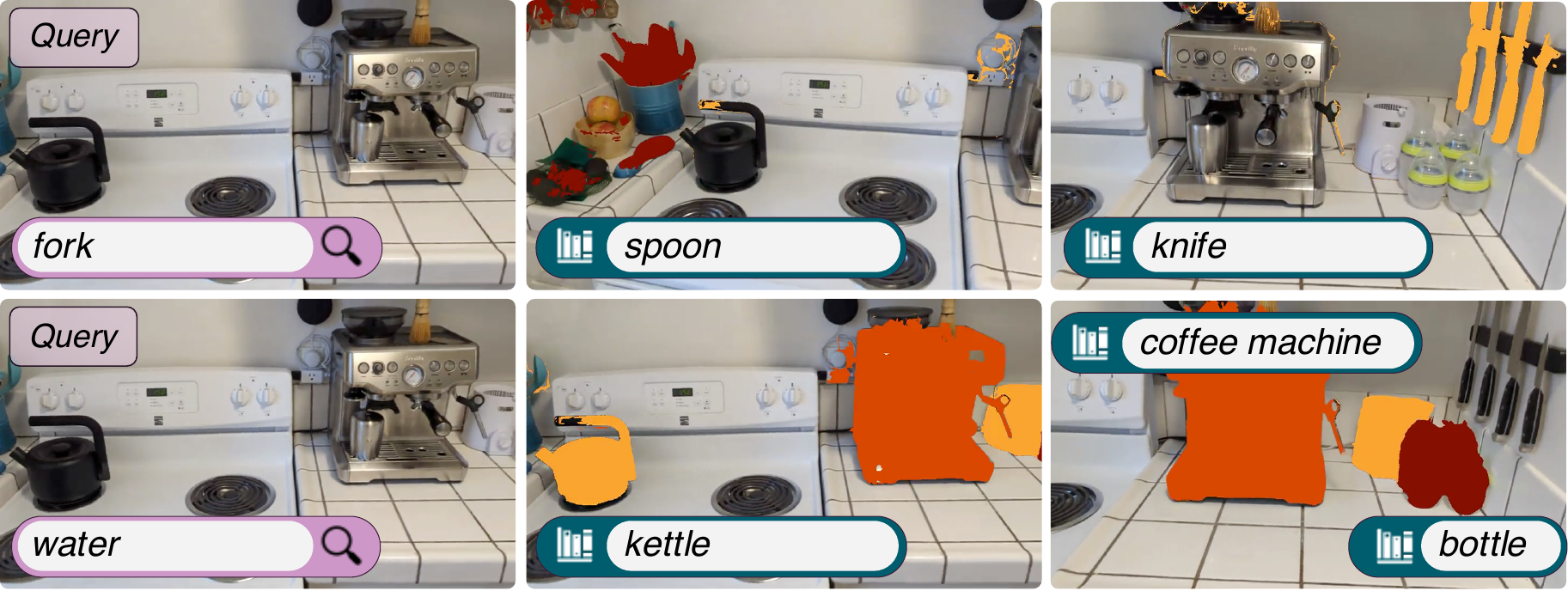}
    \caption{\algname resolves semantic ambiguity in object localization from natural-language queries. In the top row, although the queried object ``fork" does not exist in the scene, \algname localizes relevant objects in the scene, and more importantly, provides the semantic label of each of these objects. Here, \algname identifies a \emph{spoon} and a \emph{knife} and notes that the spoon (not the knife) is more similar to the fork. Likewise, in the bottom row, given the ambiguous prompt ``water," \algname localizes a \emph{kettle}, \emph{coffee machine}, and \emph{bottle}, providing their semantic object classes along with their relative similarity to the prompt.}
    \label{fig:app_experiments_disambiguation}
\end{figure*}

\section{Semantic Disambiguation}
\label{sec:appendix_experiment_semantic_disambiguation}
As discussed earlier in this work, \algname enables semantic disambiguation, e.g., in semantic object localization. Not only does \algname provide the semantic identity of the most relevant object, \algname also identifies other pertinent objects, providing the semantic label of these objects along with their similarity to the natural-language prompt. In \Cref{fig:app_experiments_disambiguation}, we demonstrate the semantic disambiguation capabilities of \algname. In the top row, when queried for a ``fork," \algname identifies the \emph{spoon} as the object that is most similar to the query. In addition, \algname localizes the \emph{knife}, identifying it as being relevant, although less similar, compared to the spoon. 
Likewise, in the bottom row, when prompted with the ambiguous query ``water," \algname localizes the \emph{kettle}, noting its lower similarity compared to a \emph{coffee machine}, which \algname also localizes in the scene. Lastly, \algname identifies the \emph{bottle} as the most similar object to the query, resolving the ambiguous prompt.

\end{appendices}

\end{document}